\documentclass[journal, a4paper]{IEEEtran}
 
\IEEEoverridecommandlockouts
\usepackage{cite}
\usepackage{amsmath,amssymb,amsfonts}
\usepackage{graphicx}
\usepackage{textcomp}
\usepackage{xcolor}
\usepackage{dsfont}
\usepackage{tabulary,booktabs}
\usepackage{multicol}
\usepackage{enumitem}

\usepackage{hyperref}
\usepackage{environ}
\usepackage{float} 
\usepackage{makecell}  
\usepackage{booktabs}
\usepackage{mathtools}
\usepackage{cancel}
\usepackage{verbatim}
\usepackage{dsfont}
\usepackage{subfigure}
\usepackage{bm}
\usepackage{threeparttable}
\usepackage{placeins}
 
\usepackage[ruled,linesnumbered]{algorithm2e}
\usepackage{float}
\usepackage{subcaption}
\usepackage{amssymb}

\newcommand{\OfaSet}{\ensuremath{\Gamma_x}}  
\newcommand{\AdaSet}{\ensuremath{\Gamma_\theta}}

\newcommand{\AdaOfaSet}{\ensuremath{\Gamma_{\tilde{x}}}}

\newcommand{\Acc}{\ensuremath{F_{AE}(\tilde{x})}}

\newcommand{\AccConstr}{\ensuremath{\overline{A}}}
\newcommand{\Params}{\ensuremath{F_P(\tilde{x})}}

\newcommand{\ParamsConstr}{\ensuremath{\overline{P}}}

\newcommand{\Cost}{\ensuremath{F_C(\tilde{x})}}

\newcommand{\TechConstr}{\ensuremath{\mathcal{TC}}}
\newcommand{\FuncConstr}{\ensuremath{\mathcal{FC}}}
\newcommand{\Prior}{\ensuremath{\beta}}
\newcommand{\EvalHyper}{\ensuremath{\delta}}

\newcommand{\params}{$F_P$}
\newcommand{\mem}{$F_{ME}$}
\newcommand{\macs}{$F_M$}
\newcommand{\flops}{$F_F$}
\newcommand{\lat}{$F_L$}
\newcommand{\energy}{$F_{E}$}
\newcommand{\edp}{$F_{EP}$}
\newcommand{\area}{$F_{AR}$}
\newcommand{\acc}{$F_A$}
\newcommand{\aia}{$F_{AIA}$}
\newcommand{\act}{$F_{AS}$}
\newcommand{\rob}{$F_R$}
\newcommand{\auc}{$F_{AC}$}
\newcommand{\ap}{$F_{AP}$}
\newcommand{\iou}{$F_{IU}$}
\newcommand{\rec}{$F_{RE}$}
\newcommand{\hit}{$F_H$}
\newcommand{\dg}{$F_{DG}$}
\newcommand{\fscore}{$F_{F1}$}
\newcommand{\rouge}{$F_{RO}$}
\newcommand{\fps}{$F_{FS}$}
\newcommand{\thr}{$F_T$}
\newcommand{\traintime}{$F_{TT}$}
\newcommand{\cachesize}{$F_{CS}$}
\newcommand{\paramsconstr}{$\overline{P}$}
\newcommand{\macsconstr}{$\overline{M}$}
\newcommand{\actconstr}{$\overline{AS}$}
\newcommand{\energyconstr}{$\overline{E}$}
\newcommand{\latencyconstr}{$\overline{L}$}
\newcommand{\flopsconstr}{$\overline{F}$}
\newcommand{\memconstr}{$\overline{ME}$}
\makeatletter
\newcommand\newtag[2]{#1\def\@currentlabel{#1}\label{#2}}
\makeatother

\def\BibTeX{{\rm B\kern-.05em{\sc i\kern-.025em b}\kern-.08em
    T\kern-.1667em\lower.7ex\hbox{E}\kern-.125emX}}
    
\begin{document}

\title{HERCULES: Hardware-Efficient, Robust, Continual Learning Neural Architecture Search}

\author{\IEEEauthorblockN{Matteo Gambella, Fabrizio Pittorino, and Manuel Roveri}\\ 
\IEEEauthorblockA{Dipartimento di Elettronica, Informazione e Bioingegneria\\
Politecnico di Milano, Milan, Italy\\
Email: \{matteo.gambella, fabrizio.pittorino, manuel.roveri\}@polimi.it} 
}

\maketitle
 

\begin{abstract}
Neural Architecture Search (NAS) has emerged as a powerful framework for automatically discovering neural architectures that balance accuracy and efficiency. However, as AI transitions from static benchmarks to real-world deployment, the traditional focus on hardware-aware efficiency is no longer sufficient. We observe that modern NAS methods, especially those that target edge AI, are evolving to address a triple objective: Efficiency, Robustness, and Continual Learning. While efficiency ensures feasibility in resource-constrained environments, robustness guarantees reliability under environmental variabilities, and continual learning enables adaptation to sequential tasks without catastrophic forgetting.
We propose a taxonomy of NAS approaches through this triple lens, distinguishing between methods targeting resource optimization, environmental resilience, and architectural plasticity. This unified perspective reveals that these axes, though often studied in isolation, are mutually reinforcing. Building on this taxonomy, we map the current landscape of these NAS methods into a new framework called Hardware-Efficient, Robust, and ContinUal LEarning Search (HERCULES). We define the desiderata, the twelve labours of HERCULES, addressing the non-trivial challenge of balancing an adequate search-space exploration with the immense computational costs of a multi-objective NAS, accounting for these crucial objectives of current AI systems. By identifying critical gaps in existing research, this survey outlines a roadmap toward integrated algorithmic, architectural, and hardware-software co-design for truly deployable, lifelong-learning AI systems.
 
\end{abstract}
 
 
\section{Introduction}
 
Deep Neural Networks (NNs) have become the de facto standard across diverse domains, including computer vision, natural language processing, and time-series forecasting \cite{TCN}. To bypass the arduous and manual process of architecture design, Neural Architecture Search (NAS) offers a principled framework for automated discovery \cite{nas_survey_components}. A prominent branch of this research is Hardware-aware NAS (HW-NAS) \cite{benmeziane2021ijcai}, which targets the critical tension between accuracy on a given data distribution and hardware \textbf{efficiency} (e.g., memory and computational demand). This trade-off is essential, as the state-of-the-art performance of modern DNNs often necessitates massive parameter counts and extreme computational complexity \cite{scalinglaw}.
 
However, accuracy on i.i.d. data and efficiency alone are not sufficient. Once deployed, an AI system must also ensure \textbf{robustness} \cite{flatnas}, that is, the ability of a model to maintain reliable performance under diverse sources of variability such as adversarial perturbations and hardware-induced noise. Furthermore, as models are increasingly deployed in dynamic environments, a third requirement emerges: \textbf{Continual Learning (CL)}, the ability to incrementally adapt to a stream of new incoming tasks \cite{continual-survey}. 
Together, efficiency, robustness, and continual learning define three orthogonal optimization directions that modern NAS must jointly consider.
 
While these three research directions are complementary, they remain distinct challenges: enhancing efficiency does not inherently ensure robustness, nor does it equip a model to navigate sequential task shifts \cite{flatnas}. Rather, they represent the three essential pillars of AI. Efficiency enables feasibility and scalability within resource-constrained environments \cite{tinymloverview}; Robustness provides reliability and trust amidst uncertainty and varying data distributions \cite{empiricalstudyrobustness}; and Continual Learning manages the vital balance between plasticity (the capacity to integrate new knowledge) and stability (the preservation of previously acquired knowledge), ensuring sustained performance over time \cite{continual-survey}.

Despite their importance, robustness, efficiency, and continual learning have often been studied in isolation in current NAS methods. Existing surveys about NAS algorithms extensively cover algorithmic strategies \cite{elsken2019nas, ren2021nas, sekanina2021nas, nasframework, advancesnas}, hardware-aware design \cite{chitty2022nas, benmeziane2021ijcai, sekanina2021nas}, and individual treatments of robustness \cite{empiricalstudyrobustness, reviewNASIMC} or continual learning \cite{continual-survey}. Yet, none provide a unified perspective that explicitly treats these three pillars as equally fundamental and orthogonal objectives in NAS research. Table \ref{tab:survey_comparison} summarizes the coverage of existing surveys along the three axes of efficiency, robustness, and continual learning, highlighting this gap.
 
In this survey, we address this gap by exploring a central research question: \textit{Is it possible to automate the design of neural network architectures that maintain performance reliability under hardware noise and data shifts while respecting strict energy budgets and evolving task requirements?}
 
In this direction, we propose a taxonomy of NAS approaches through the triple lens of robustness, efficiency, and continual learning, identifying how different algorithmic, architectural, and hardware-level strategies contribute to progress along these three axes.
 
To answer the research question, we define a new framework targeting these axes jointly that we name \textbf{HERCULES (Hardware-Efficient, Robust, and Continual LEarning Search)}. Mirroring the twelve labors of its mythological namesake, HERCULES represents the daunting task of navigating an immense search space to find architectures capable of surviving the challenges of the deployment of DNNs, ranging from hardware constraints to adversarial attacks and shifting data streams. Additionally, the framework assigns a central role to the aspect of adaptivity of DNNs (e.g., adaptation to the difficulty of the input), since it can bring benefits over all the three axes.
 
The structure of our survey is the following. The key concepts of NAS are presented in Section \ref{sec:nas_background}. 
The NAS solutions for efficient neural networks are introduced in Section \ref{sec: efficient_nas}, NAS methods for robust NNs are discussed in Section \ref{sec:robust_nas}, while NAS solutions for Continual Learning are commented in Section \ref{sec:nas_continual}.
Then, Section \ref{sec:robust_efficient_nas} explores solutions that jointly optimize the three introduced pillars. We highlight that few NAS works optimize Continual Learning in isolation.
The definition of HERCULES is introduced in Section \ref{subsec:hercules_framework} while its desiderata are commented in Section \ref{sec:desiderata}. 
Future directions and conclusions are presented in Section \ref{sec:future_directions} and Section \ref{sec:conclusion}. Lastly, given the wide scope of this survey, we summarized the datasets used in the examined works, divided according to the task, in Table \ref{tab:datasets_grouped} and the metrics used in the examined works in Table \ref{tab:metrics}.

\section{Neural Architecture Search}
\label{sec:nas_background}
 
Neural Architecture Search (NAS) methods can be categorized along three key dimensions~\cite{nas_survey_components}.
 
The search space defines how architectures are represented. Entire-structured spaces model networks layer by layer, while cell-based spaces, inspired by hand-crafted designs like MobileNets~\cite{mobilenetv3}, group layers into reusable blocks or cells \cite{liu_darts_2019}. Extensions include hierarchical structures and morphism-based transformations.
 
The search strategy determines how the space is explored. Common approaches include reinforcement learning~\cite{rlsurvey}, gradient-based methods~\cite{liu_darts_2019, zodarts+, zodarts++}, diffusion models \cite{diffusionNAS}, Large Language Models \cite{zeinaty2025}, and evolutionary algorithms~\cite{gambella2023edanas, nachos}. Genetic strategies, especially NSGA-II~\cite{nsganetv2,cnas}, are widely adopted for multi-objective optimization.
 
The performance estimation strategy aims at evaluating NNs without full training, which is computationally infeasible. A common approach is weight sharing, exemplified by Once-For-All (OFA)\cite{ofa}, which trains a supernet once and evaluates candidates by fine-tuning its weights. Another strategy uses surrogate models like Gaussian Processes or Radial Basis Functions. For example, MSuNAS\cite{nsganetv2} employs adaptive-switching to select the most accurate predictor among four surrogates using Kendall’s Tau correlation in each NAS iteration. Zero-shot NAS methods~\cite{zeroshot, Mellor2021NASWOT} further reduces cost by relying on training-free proxies correlated with final accuracy.
 
Standard Neural Architecture Search (NAS) focuses primarily on a single-objective optimization: finding the architecture that achieves the highest accuracy on a target task and a dataset. The problem formulation is defined as follows:
 
\begin{align}
\label{eq:basenas_problem}
    \text{maximize }  &  F_A(x)    
    \\
    \text{s. t. }  & x \in \Gamma_{x} \nonumber
\end{align}
 
where $x$ and $\Gamma_{x}$ represent a candidate NN architecture and the corresponding search space, respectively, and $F_A(x)$ is the accuracy of the NN.
 
Overall, a NAS framework takes as input a dataset, evaluation hyperparameters \EvalHyper, a family of neural networks parameterized by an architectural set \OfaSet{} defining the Search Space, and outputs the network, found by using the mono-objective Search Strategy, with the highest accuracy in the Search Space.
 
This traditional formulation acts as the baseline for the multi-objective approaches discussed in the following sections, which expand this objective to include hardware efficiency, robustness, and continual learning capabilities.
 
\begin{table*}[t]
\centering
\caption{Comparison of existing NAS surveys along the three AI pillars: efficiency, robustness, and continual learning.
\checkmark~indicates the axis is explicitly and thoroughly addressed;
$\circ$~indicates partial or implicit coverage;
$\times$~indicates the axis is not addressed. }
\label{tab:survey_comparison}
\renewcommand{\arraystretch}{1.25}
\begin{tabular}{lcccc}
\toprule
\textbf{Survey} & Topics & \textbf{Efficiency} & \textbf{Robustness} & \textbf{Continual Learning}  \\
\midrule
Elsken et al.~\cite{elsken2019nas} & \makecell[t]{The three basic components: search space, search strategy, \\ and performance estimation strategy}                        & $\circ$  &  $\times$   & $\times$   \\
Salmani et al.~\cite{salmanipouravval2025systematic} & \makecell[t]{The three basic components: search space, search strategy, \\ and performance estimation strategy}                        & $\circ$  &  $\times$   & $\times$   \\
Ren et al.~\cite{ren2021nas}    &   The methodological challenges of NAS and their solutions                  & $\circ$ &   $\times$   & $\times$   \\

Liu et al. \cite{survey-CE-NAS} &
The challenge of the computational efficiency of NAS methods & $\times$ & $\times$ & $\times$ \\
Liu et al. \cite{liu2023enas_survey} & The evolutionary NAS methods discussed on their core components & $\circ$  &  $\times$   & $\times$ \\
Li et al. \cite{zeroshot} & Challenges, solutions, and opportunities of Zero-Shot NAS & \checkmark &  $\times$   & $\times$ \\
Zhu et al. \cite{surveyfederatednas} & \makecell[t]{The NAS methods for Federated Learning discussed on their search \\ strategies and offline/online implementations} & $\circ$ &   $\times$   & $\times$ \\
Benmeziane et al.~\cite{benmeziane2021ijcai}   & \makecell[t]{Hardware-aware NAS on fixed hardware and hardware-software \\ codesign divided according to the methodologies}          & \checkmark &  $\times$   & $\times$    \\
Chitty-Venkata \& Somani~\cite{chitty2022nas} & \makecell[t]{Hardware-aware NAS: fixed hardware and hardware-software \\ codesign divided according to the targeted platforms} & \checkmark &  $\times$   & $\times$    \\
Sekanina~\cite{sekanina2021nas}   & \makecell[t]{Hardware-aware NAS on fixed hardware and hardware-software \\ codesign divided according to the methodologies}                      & \checkmark & $\times$   & $\times$   \\
Poyser \& Breckon~\cite{nasframework} & \makecell[t]{NAS approaches with respect to image classification, object detection, \\ and image segmentation}                          & $\circ$    & $\times$   & $\times$   \\
Wang \& Zhu~\cite{advancesnas}         & \makecell[t]{Delves into the multifaceted aspects of NAS: \\ recent advances, applications, tools, benchmarks and research directions}                   & $\circ$    & $\times$   & $\times$    \\
Devaguptapu et al.~\cite{empiricalstudyrobustness} & Adversarial robustness of NAS-based architectures & $\times$   & \checkmark & $\times$    \\
Krestinskaya et al.~\cite{reviewNASIMC}    & \makecell[t]{NAS methods for efficient deployment \\ on IMC hardware}              & \checkmark & \checkmark & $\times$    \\
Shahawy et al.~\cite{surveynascontinual} & NAS methods for continual learning                                 & $\times$   & $\times$   & \checkmark  \\
\midrule
\textbf{Our survey}   & NAS methods optimizing efficiency, robustness, and continual learning                                    & \checkmark & \checkmark & \checkmark  \\
\bottomrule
\end{tabular}
\end{table*}

\begin{table}[ht!]
    \centering
    \scriptsize 
    \renewcommand{\arraystretch}{1.2}
    \begin{tabular}{|l|p{5.5cm}|}
        \hline
        \textbf{Dataset} & \textbf{Description} \\
        \hline
        
        \multicolumn{2}{|c|}{\textbf{Image Classification}} \\ \hline
        CIFAR-10/100[-C]  & Tiny images (10/100 classes). -C refers to the corrupted version. \\
        MNIST & Handwritten digits (10 classes). \\
        Fashion MNIST & Clothing articles (10 classes). \\
        SVHN & Street View House Numbers (10 classes). \\
        ImageNet & Object recognition (1k classes). \\
        ImgNet16 & ImageNet16-120: Downsampled variant (120 classes) with image size 16x16. \\
        \makecell[l]{TinyImgNet} & TinyImageNet: Subset of ImageNet (200 classes). \\
        Caltech-X & Object recognition dataset (X object classes, including one special). \\
        CINIC-10 & A dataset derived from CIFAR-10 and ImageNet (10 classes). \\
        Imagenette & Easy ImageNet subset (10 classes). \\
        MedMNIST & Medical images (Classes vary). \\
        \makecell[l]{VWW} & Visual WakeWords: Person detection for microcontrollers (2 classes). \\
        \hline
        
        \multicolumn{2}{|c|}{\textbf{Robustness \& OOD}} \\ \hline
        CIFAR-10/100-C & Corrupted CIFAR for robustness benchmarking. \\
        ImageNetV2 & Test set distribution shift (1k classes). \\
        ImageNetA & Natural adversarial examples (200 classes). \\
        NICO & Non-I.I.D. with Contexts (19 classes). \\
        \hline
        
        \multicolumn{2}{|c|}{\textbf{Continual Learning}} \\ \hline
        P-MNIST & Permutated MNIST: Pixels shuffled by a fixed matrix (used for Domain IL). \\
        R-MNIST & Rotated MNIST: Digits rotated by fixed angles. (used for Domain IL) \\
        
        \hline
        
        \multicolumn{2}{|c|}{\textbf{Detection \& Segmentation}} \\ \hline
        COCO & Common Objects in Context (80 categories). \\
        CityScapes & Urban street scenes (30 classes). \\
        OoD-FP & OOD / False Positive tests. \\
        \hline
        
        \multicolumn{2}{|c|}{\textbf{Vision-Language \& Grounding}} \\ \hline
        Flickr30K & 31k images, 5 captions each. \\
        RefCOCO/+/g & Visual grounding queries on COCO images. \\
        \hline
        
        \multicolumn{2}{|c|}{\textbf{Visual QA \& VLM}} \\ \hline
        VQA-v2 & QA pairs for COCO (3k answers). \\
        JA-VG-VQA & Japanese Visual Genome VQA. \\
        JA-VG-VQA-500 & Subset of JA-VG-VQA. \\
        JA-VLM-Bench & Japanese VLM In-the-Wild. \\
        \hline
        
        \multicolumn{2}{|c|}{\textbf{Audio \& Time-Series}} \\ \hline
        MI ECG & Myocardial Infarction from ECG (2 classes). \\
        wHAR & Wearable sensor data (7 classes). \\
        MIMII & Malfunctioning machine sound anomaly detection. \\
        \makecell[l]{Keyword Spotting \\ (KWS)} & Speech command recognition. \\
        SpeechCmd & Google Speech Commands (35 classes). \\
        MSWC & Multilingual Spoken Words ($>$300k classes). \\
        LibriSpeech-SAM & Large corpus of read English speech. \\
        \hline
        
        \multicolumn{2}{|c|}{\textbf{NLP \& Reasoning}} \\ \hline
        PTB & Penn Treebank. \\
        GSM8k & Grade School Math problems. \\
        MGSM-JA & Multilingual GSM (Japanese). \\
        JP-LMEH & Japanese Healthcare LM evaluation. \\ 
        Math & A set of Math problems. \\
        CSense & A set of common-sense reasoning tasks.
        \\
        \hline
        
        \multicolumn{2}{|c|}{\textbf{Re-Identification \& Action}} \\ \hline
        SYSU-MM01 & Visible-Thermal Re-ID (395 ids). \\
        RegDB & Visible-Thermal Re-ID (412 ids). Two modes: Visible to Infrared (V-I) or vice versa (I-V). \\
        NTU RGB-D & 3D human activity (60/120 classes). \\
        EgoGesture & Egocentric hand gestures (83 classes). \\
        \hline
        
        \multicolumn{2}{|c|}{\textbf{Miscellaneous / Multimodal}} \\ \hline
        MM-IMDB & Posters + plots for genre (23 classes). \\
        AV-MNIST & Audio-Visual digits (10 classes). \\
        Beauty/Cell/Groc & Amazon reviews (5 rating classes). \\
        FACET & CV fairness across demographics. \\
        Chars74k & Natural image characters (74 classes). \\
        \hline
    \end{tabular}
    
    \caption{Datasets grouped by Task} 
    \label{tab:datasets_grouped}
\end{table}

\begin{table}[t!]
    \centering
    \small
    \renewcommand{\arraystretch}{1.2}
    \begin{tabular}{|l|p{5.5cm}|}
        \hline
        \textbf{Metric} & \textbf{Description} \\
        \hline
        
        \multicolumn{2}{|c|}{\textbf{General Performance}} \\ \hline
        \acc & Percentage of correct predictions out of total samples. \\
        \auc & Area Under the ROC Curve; measures separability between classes. \\
        \rob{($\Delta$)} & Accuracy maintained under perturbation defined by $\Delta$. \\
        \hline
 
        \multicolumn{2}{|c|}{\textbf{Efficiency \& Hardware}} \\ \hline
        \params & Total count of learnable weights (impacts storage). \\
        \mem & RAM/VRAM usage during inference or training. \\
        \macs & Multiply-Accumulate operations; measures computational complexity. \\
        \flops & Floating Point Operations; roughly $2 \times$ MACs. \\
        \fps & Frame per seconds. \\
        \thr & Throughput \\ 
        \act & Activation sizes \\ 
        \lat & Time taken to process a single input (inference time). \\
        \energy & Power consumption integrated over time for an operation. \\
        \edp & Energy-Delay Product ($Energy \times Latency$); balances speed and efficiency. \\
         \area & Physical silicon area required for the circuit/accelerator. \\
         \traintime & Time to train the NN. \\
         \cachesize & KV cache size. \\
        \hline
 
        \multicolumn{2}{|c|}{\textbf{Object Detection \& Segmentation}} \\ \hline
        \ap & Mean Average Precision; standard metric for object detection quality. \\
        \iou & Mean Intersection over Union; measures pixel-wise overlap in segmentation. \\
        \hline
 
        \multicolumn{2}{|c|}{\textbf{Retrieval \& Recommendation}} \\ \hline
        \rec(X) & Proportion of relevant items retrieved in the top $X$ results. \\
        \hit(X) & 1 if the target is present in top $X$ results, 0 otherwise. \\
        \dg(X) & Normalized Discounted Cumulative Gain; rewards correct ranking order. \\
        \hline
        
        \multicolumn{2}{|c|}{\textbf{NLP \& Fine-Grained Classification}} \\ \hline
        \fscore & Weighted F1-score; harmonic mean of precision/recall weighted by class support. \\
        \rouge & Measures overlap (Longest Common Subsequence) in text summarization. \\
        \hline
    \end{tabular}
    \caption{Neural Network Evaluation Metrics}
    \label{tab:metrics}
\end{table}

\section{NAS for efficient NN architectures}
\label{sec: efficient_nas}
 
\subsection{Introduction}
\label{subsec:background_efficiency}
 
Hardware-Aware NAS methods aim to design architectures that 
optimize not only accuracy but also hardware-related metrics 
such as memory footprint, computational cost, and energy 
consumption~\cite{benmeziane2021ijcai}, marking a paradigm 
shift from accuracy-centric search toward multi-objective 
optimization. Three families of approaches have been proposed: 
(i)~methods searching for compressed static architectures, 
(ii)~algorithms exploring dynamic neural networks that adapt 
computation to input complexity, and (iii)~hardware--software 
co-design techniques jointly optimizing architecture and 
accelerator.

TinyML~\cite{tinymloverview} targets low-power microcontrollers, 
embedded devices, and IoT nodes via three compression techniques: \emph{Knowledge 
Distillation}~\cite{abbasi2020modeling}, which transfers 
knowledge from a large teacher to a smaller student network; 
\emph{Pruning}~\cite{han2015learning}, which removes redundant 
weights or channels; and \emph{Quantization}~\cite{han2015deep}, 
which converts floating-point weights into lower-precision 
integer formats~\cite{velasquez2025tinyml, cnas, 
perceptionmodel}.

Dynamic Neural Networks~\cite{surveydynamicNN} reduce inference cost by 
adapting computation to each input. Key classes include: 
Early-Exit Neural Networks (EENNs)~\cite{scardapane_why_2020, 
dynamax}, which terminate inference early when confidence is 
sufficient; Mixture-of-Experts (MoE)~\cite{jacobs1991moe}, 
which route inputs through specialized subnetworks; Capsule 
Networks~\cite{capsnets}, which encode pose and presence via 
dynamic routing between capsules; and Multi-Branch 
Networks~\cite{odena2017changingmodelbehaviortesttime, 
branchLLM}, which select per-input computation paths.
 
\subsection{Overview of the NAS methods}

Different from the problem formulation of the standard NAS, introduced in Eq. \ref{eq:basenas_problem}, the multi-objective problem of a Hardware-Aware NAS is built on a refined search space, integrates the cost functions, and applies constraints if needed. The problem formulation is the following:
 
\begin{align}
\label{eq:adacnas_problem}
    \text{minimize }  &\mathcal{G} \left 
    ( F_A(\tilde{x}),
    F_C^1(\tilde{x}),
    ..,
    F_C^N(\tilde{x}),
    \right )
    \\
    \text{s. t. }  & \tilde{x} \in \Gamma_{\tilde{x}} \nonumber
    \\
    & F_A(\tilde{x}) > \bar{A}
    \nonumber \\
    & F_C^i < \bar{C_i} , \quad \forall i = 1, \dots, N 
    \nonumber
\end{align}
where $\mathcal{G}$ is a multi-objective optimization function, $\tilde{x}$ and $\Gamma_{\tilde{x}}$ represent a candidate \FuncConstr-compliant NN architecture 
and the corresponding search space,
respectively, $F_A(\tilde{x})$ is the accuracy under no perturbations of the environment and $\bar{A}$ is the related constraint, $F_C^i(\widetilde{x})$ is a cost function of $\tilde{x}$ and $\bar{C_i}$ is the related constraint, belonging to the set of technological constraints \TechConstr.

Note that $F_A(\tilde{x})$ appears both as a component of the 
multi-objective function $\mathcal{G}$ and as a hard constraint. 
This is intentional: $\mathcal{G}$ drives the Pareto-optimal 
trade-off among all objectives, while the constraint $F_A > \bar{A}$ 
enforces a minimum acceptable accuracy floor, ensuring that no 
solution sacrifices task performance beyond a deployment-defined 
threshold regardless of its efficiency gains.
 
We can refer to the search space describing only the NN architecture as $\Gamma_x$ while we define as $\Gamma_{\tilde{x}}$ a refined search space that incorporates other aspects such as the dynamic scheme of the NN (e.g., early exits) or the compiler mapping for the hardware co-optimization. This is the search space for (ii) and (iii) methodologies. From an architecture perspective of the NAS, we can define an \textbf{Arch Adapter} module that enhances the base network architectures sampled from $\Gamma_x$ with dynamic capabilities specified by a configuration sampled from \AdaSet, and defines the novel NAS search space set $\Gamma_{\tilde{x}}$. For instance, \AdaSet{} may contain the placement and thresholds of early exit classifiers \cite{gambella2023edanas, nachos, aebnas}.
 
Another novel addition regards the concept of constraint. Many Hardware-Aware NAS don't simply optimize some metrics but explicitly delineate which architectures are considered feasible solutions for the NAS. The set of constraints extends the formalization introduced by CNAS \cite{cnas} and can be formalized as follows:
    \begin{itemize}
      \item \textbf{Hardware specifications (e.g., compiler mappings, CM)}: Describe the target hardware accelerator.
      \item \textbf{Technological constraints (\TechConstr)}: Reflect computational and memory limitations.
      \item \textbf{Functional constraints (\FuncConstr)}: Restrict allowable operations, e.g., polynomial operations for encrypted inference or hardware-compliant operations for some specific hardware.
    \end{itemize}
 
The output is a set of \textbf{Pareto-optimal neural networks}, which may be classified as:
\begin{itemize}
  \item \textbf{Dynamic}, if dynamic behavior is incorporated (i.e., \( \AdaSet \neq \emptyset \)). We refer to dynamic architectures as $\tilde{x}$.
  \item \textbf{Static}, if no dynamic behavior is included. We refer to static architectures as $x$.
\end{itemize}

We summarized the metrics used in the examined works, divided according to their macrocategory, in Table \ref{tab:metrics}. We highlight that the efficiency and hardware metrics can be the cost functions \Cost{} of our work, while the other metrics refer to the accuracy and robustness of the designed models.
 
We emphasize the presence of hardware-agnostic metrics such as the MACs, and others addressing hardware-dependent metrics such as the latency, often by using a Look Up Table (LUT) containing each layer cost on the target hardware to speed up evaluation \cite{proxylessnas}. This aligns with the observation that some Hardware-NAS methods are not tailored to specific hardware.

\begin{table*}[h!]
\caption{Summary of NAS methods with model compression. Objectives refer to the metrics optimized by the NAS under the given constraints (annotated with overline). Search Cost is in GPU days/hours. Results show exact metrics. Datasets reported with only the right bracket are the ones used for search, while datasets reported inside both brackets are only used for evaluation.}
\label{tab:compression_nas_methods}
\centering
\resizebox{\textwidth}{!}{%
\begin{tabular}{l l l l l}
\toprule
\textbf{Model} & \textbf{Obj / Constr} & \textbf{Method} & \textbf{Search Cost} & \textbf{Results} \\
\midrule
MSuNAS \cite{lu2020nsganetv2}
    & \acc, \params, \macs, \lat
    & Surrogate-assisted Genetic Search
    & $\sim$1 GPU-d*
    & \makecell[l]{ImageNet) \acc: 80.4\%, \params: 8.7M, \macs: 593M \\ \lat: 16.7 ms (CPU), \lat: 73 ms (GPU) \\
      \textit{Device: Intel i7-8700K (CPU), GTX 1080Ti (GPU)}}
\\
\midrule
CNAS \cite{cnas}
    & \makecell[l]{\acc, \params, \macs, \act{} \\ / \paramsconstr, \macsconstr, \actconstr}
    & Constrained Genetic Search
    & $\sim$1 GPU-d*
    & \makecell[l]{CIFAR-10) \acc: 89.9\%, \params: 2.19M, \macs: 6.85M, \\ \act: 0.23M}
\\
\midrule
RC-DARTS \cite{jin2019rc}
    & \acc, \params, \flops{} / \paramsconstr, \flopsconstr
    & Constrained Gradient Search
    & $\sim$1 GPU-d
    & \makecell[l]{CIFAR-10) \Acc: 97.2\%, \params: 3.3M \\ ImageNet) \Acc: 74.9\%, \Params: 4.9M, \flops: 590M}
\\
\midrule
HardCore-NAS \cite{hardcorenas}
    & \acc, \lat{} / \latencyconstr
    & Constrained Gradient Search
    & $\sim$400 GPU-h
    & \makecell[l]{ImageNet) \acc: 78.0\%, \lat: 61 ms \\
      \textit{Device: Intel Xeon (CPU)}}
\\
\midrule
ZO-DARTS++ \cite{zodarts++}
    & \acc, \params{} / \paramsconstr
    & Constrained Gradient Search
    & $\sim$0.4 GPU-h
    & MedMNIST) \acc: 84.00\%, \params: 0.83M
\\
\midrule
FBNet \cite{wu2019fbnet}
    & \acc, \lat
    & Gradient Search with LUT
    & $\sim$9 GPU-d
    & \makecell[l]{ImageNet) \acc: 74.1\%, \flops: 295M, \lat: 23.1ms \\
      \textit{Device: Samsung S8 Phone}}
\\
\midrule
ProxylessNAS \cite{cai2019proxylessnas}
    & \acc, \lat
    & Path-level Pruning
    & $\sim$200 GPU-h
    & \makecell[l]{CIFAR-10) \acc: 97.92\%, \params: 5.7M \\ ImageNet) \acc: 74.6\%, \lat: 78ms \\
      \textit{Device: Mobile}}
\\
\midrule
OFA \cite{ofa}
    & \acc{} / \latencyconstr, \energyconstr, \macsconstr
    & \makecell[l]{Sampling from the supernet \\ respecting constraints}
    & $\sim$40 GPU-h\textsuperscript{*}
    & ImageNet) \acc: 80.0\%, \macs: 595M
\\
\midrule
DNAS \cite{wu2018mixed}
    & \acc, \macs
    & Mixed-precision Gradient Search
    & $\sim$40 GPU-h
    & \makecell[l]{CIFAR-10) \acc: 95.07\%, \params: $\times$12.5\% compression \\ (w.r.t. ResNet110 FP) \\ ImageNet) \acc: 74.12\%, \params: $\times$87.4\% compression \\ (w.r.t. ResNet34 FP)}
\\
\midrule
APQ \cite{wang2020apq}
    & \acc, \lat, \energy
    & Joint Arch+Prun+Quant Search
    & $\sim$40 GPU-h\textsuperscript{*}
    & ImageNet) \acc: 75.1\%, \lat: 12.17ms, \energy: 14.14 mJ
\\
\midrule
HAPFNAS \cite{hapfnas}
    & \acc / \flopsconstr
    & \makecell[tl]{Evolutionary Search on Federated NAS \\ with a budget of FLOPS on the clients}
    & $\sim$ \makecell[l]{20 GPU-min/client + \\ 3.5--7.6 GPU-h (supernet)}
    & \makecell[tl]{CIFAR-10) \acc: 85.38\%, CIFAR-100) \acc: 64.73\%, \\ TinyImgNet) \acc: 46.47\%, $N$=12 clients}
\\
\midrule
MicroNets \cite{banbury2021micronets}
    & \acc{} / \latencyconstr, \memconstr
    & Differentiable Search on Quantized NNs
    & -
    & \makecell[l]{VWW) \acc: 85.0\%, \lat: 1.06s \\ KWS) \acc: 96.5\%, \lat: 0.19s \\ AD) \auc: 97.28\%, real-time detection \\
      \textit{Device: Medium MCU (VWW), Small MCU (KWS, AD)}}
\\
\midrule
LoNAS \cite{munoz-etal-2024-lonas}
    & \acc{}, \flops{}
    & \makecell[tl]{Genetic Search on LLMs with width \\ pruning via LoRA adapters}
    & $\sim$1--2 GPU-d + supernet
    & \makecell[tl]{GLUE) \acc: 83.06\%, \flops: 8.00G \\ CSense) \acc: 65.20\%, \flops: 1.40T, \params: 5.60B \\ Math) \acc: 37.20\%, \flops: 1.50G, \params: 6.10B}
\\
\midrule
LoNAS-2 \cite{compressingllm}
    & \acc{}, \flops{}
    & \makecell[tl]{Genetic Search on LLMs with width \\ and depth pruning via LoRA adapters}
    & -
    & CSense + Math) \acc: 44.49\%, \lat: 86.22ms, \params: 6.06B
\\
\midrule
Composer \cite{composer}
    & \makecell[tl]{\acc{}, \lat{}, \traintime, \\ \cachesize}
    & \makecell[tl]{Bayesian Search over ratio and \\ interleaving patterns of primitives \\ within Transformers \& synthetic \\ search (MAD) dataset}
    & $\sim$5--30 GPU-h
    & \makecell[tl]{MAD) 61.6\% (CSense), TrainTime: $\approx$2.7s/step, \\ \lat: $\approx$225s, KV cache: $\approx$0.3GiB \\ (est. from figures, 1B, bs=1, seq=8K)}
\\
\bottomrule
\multicolumn{5}{l}{\footnotesize \textsuperscript{*} OFA requires huge upfront training (supernet) around 1200 GPU hours on ImageNet.}
\end{tabular}%
}
\end{table*}
 
\begin{table*}[h!]
\caption{Summary of NAS methods for Hardware-Software Co-design. Objectives refer to the metrics optimized by the NAS under the given constraints (annotated with overline). Search Cost is in GPU days/hours. Results show exact metrics. Datasets reported with only the right bracket are the ones used for search, while datasets reported inside both brackets are only used for evaluation.
}
\label{tab:codesign_nas_methods}
\centering
\resizebox{\textwidth}{!}{%
\begin{tabular}{l l l l l}
\toprule
\textbf{Model} & \textbf{Obj / Constr} & \textbf{Method} & \textbf{Search Cost} & \textbf{Results} \\
\midrule
Codesign-NAS \cite{codesignnas} & 
\acc{}, \lat, \area / \AccConstr, \latencyconstr, $\overline{AR}$ & FPGA/CNN Joint Search by RL & $\sim$96 GPU-d & \makecell[tl]{ CIFAR-100)  \acc: 72.00\%, \lat: 18.5ms,  \\ \area: 133.00 $mm^2$ ; \textit{Device}: FPGA}\\  
\midrule
NAAS \cite{lin2021naas} & \edp, \energy, \lat & \makecell[l]{Arch+Map+Accel (ASIC) Evolutionary \\ Search} & $\sim$6 GPU-h* & \makecell[l]{CIFAR-10) \acc: 93.2\%, \lat: 20000 cycles, \energy: 2J, \\ \edp: 200000 J*clock cycle \\ \textit{Device: ASIC}} \\
\midrule
EDD \cite{li2020edd} & \acc, \lat, \thr & Accel./CNN Joint Gradient Search & $\sim$12 GPU-h & \makecell[l]{ImageNet) \acc: 74.7\%, \lat: 11.17ms (GPU), \\ \lat: 11.15ms (FPGA); \textit{Device: GPU, FPGA}}
\\
\midrule
DNA \cite{zhang2020dna} & \acc, \lat, \fps, \edp & \makecell[l]{Accel.(FPGA/ASIC)/CNN Joint Gradient \\ Search} & \makecell[tl]{$\sim$4.2 GPU-h (C10/100); \\ $\sim$144 GPU-h (ImgNet)} & \makecell[l]{FPGA) CIFAR-10) \acc: 96.10\%, \fps: 52.4; \\ CIFAR-100) \acc: 79.35\%, \fps: 64.3; \\ ImageNet) \acc: 75.70\%, \fps: 31.9; \\ ASIC) CIFAR-10) \acc: 96.50\%, \edp: 4990 J*clk \\ \textit{Device: FPGA / ASIC}}
\\
\midrule
AutoDNN \cite{hao2019fpga_dnn_codesign} & \acc{} / \latencyconstr, \energyconstr & FPGA/CNN Joint Search by SCD & - & \makecell[l]{AD) \iou: 59\%, \lat: 33.7 ms, \energy: 4.04 kJ \\ \textit{Device: FPGA}} \\
\midrule
NAHAS \cite{zhou2021rethinking_codesign} & \acc / \latencyconstr, $\overline{AR}$ & EdgeTPU/CNN Joint Search by PPO/RL  & \makecell[tl]{ $\sim$32 TPU-d ; \\ $\sim$1 TPU-d (w. sharing)} & \makecell[tl]{ ImageNet) \acc: 76.50\%, \lat: 350 us \\ \acc: 78.60\% ,  \lat: 2000 us (weight sharing) \\ \textit{Device: EdgeTPU}} \\
\midrule
FNAS \cite{jiang2019fpga_aware_nas} & \acc, \lat & Arch+Accel (FPGA) Analytical Search & analytical & \makecell[l]{MNIST) \acc: 98.61\%, \lat: 1.80ms \\ \textit{Device: FPGA}} \\
\midrule
NAS4RRAM \cite{yuan2021nas4rram} & \acc, \energy, \lat{} / \ParamsConstr & Arch+RRAM Evolutionary Search & - & \makecell[l]{CIFAR-10) \acc: 84.4\%, \params: 383k, \\ CIFAR-100) \acc: 53.1\%, \params: 343k \\ \textit{Device: RRAM accelerator}} \\
\bottomrule
\end{tabular}%
}
\end{table*}

\begin{table*}[h!]
\caption{Summary of NAS methods for Dynamic Neural Networks. Objectives refer to the metrics optimized by the NAS under the given constraints (annotated with overline). Search Cost is in GPU days/hours. Results show exact metrics. Datasets reported with only the right bracket are the ones used for search, while datasets reported inside both brackets are only used for evaluation.
}
\label{tab:dynamic_nas_methods}
\centering
\resizebox{\textwidth}{!}{%
\begin{tabular}{l l l l l}
\toprule
\textbf{Model} & \textbf{Obj / Constr} & \textbf{Method} & \textbf{Search Cost} & \textbf{Results} \\
\midrule
EExNAS \cite{eexnas} & \acc, \macs & Arch Bayesian Search + 1 EE block  & $\sim$0.5 GPU-d & \makecell[l]{MI ECG) \acc: 98.54\%, \mem: 15.97kB, \energy:28.19mJ, \\ wHAR) Top1: 96.77\%, \mem: 2.10kB, \energy: 0.72mJ} \\
\midrule
HADAS \cite{hadas} & \acc, \lat, \energy & \makecell[l]{Arch+EE+Accel(Edge devices) Genetic \\ Search} & $\sim$2--3 GPU-d & CIFAR-100) \acc: 93.16\%, \energy: 93.78 mJ \\
\midrule
NASEREX \cite{naserex} & \acc, \lat & EE evolutionary search & - & \makecell[l]{CIFAR-10) \acc: 85.28\%, \lat: 1.78ms; \\ CIFAR-100) \acc:49.76\%, \lat: 2.17ms; \\
CalTech-101) \acc: 78.07\%, \lat: 1.43ms; \\ CalTech-256) \acc: 35.23\%, \lat: 2.41ms} \\
\midrule
EDANAS \cite{gambella2023edanas} & \acc, \macs & \makecell[l]{Backbone Arch.+EE placement/thresholds \\ Genetic Search} & $\sim$2 GPU-d & CIFAR-10) \acc: 81.1\%, \macs: 23.52M \\
\midrule
NACHOS \cite{nachos} & \acc, \macs{} / \macsconstr & \makecell[l]{\makecell[l]{Backbone Arch.+EE placement/thresholds \\ /pooling size} \\ Constrained Genetic Search} & $\sim$2 GPU-d & \makecell[l]{SVHN) \acc: 79.96\%, \macs: 1.46M; \\ CIFAR-10) \acc: 72.65\%, \macs: 2.44M; \\ CINIC-10) \acc: 60.85\%, \macs: 2.42M, \\ Imagenette) \acc: 94.89\%, \macs: 92.34M} \\
\midrule
AEBNAS \cite{nachos} & \acc, \macs{} / \macsconstr & \makecell[l]{Backbone Arch.+EE placement/thresholds \\ /arch. Constrained Genetic Search} & - & \makecell[l]{SVHN) \acc: 84.02\%, \macs: 1.10M; \\ CIFAR-10) \acc: 74.64\%, \macs: 2.47M, \\ CIFAR-100) \acc:69.90\% \macs: 14.80M}\\
\midrule
Zniber et al. \cite{zniber2025hardwareawareneuralarchitecturesearch} & \acc, \energy, \lat{} / \energyconstr, \latencyconstr & \makecell[l]{Quantized Arch+EE Constrained Genetic \\ Search for edge accelerators} &- & \makecell[l]{CIFAR-10, INT8 bitwidth) \acc: 88.51\%, \\ \energy*\lat: 467 J*cycles; \\ CIFAR-10, INT4 bitwidth) \acc: 86.01\%, \\ \energy*\lat: 186 J*cycles} \\
\midrule
S2DNAS \cite{s2dnas} & \acc, \flops & Arch+EE(Channel) Gradient Search & - & \makecell[l]{CIFAR-10) \acc: 92.25\%, \flops: 25M; \\ CIFAR-100) \acc: 73.50\%, \flops: 39M} \\
\midrule
NASCaps \cite{nascaps} & \acc, \lat, \energy, \mem & Arch+Caps Genetic Search & $\sim$1 GPU-d & \makecell[l]{CIFAR-10) \acc: 85.99\%, \energy: 17.38mJ, \\ \lat: 1.53ms, \mem: 6319kB \\ \textit{Device: CapsAcc ASIC}} \\
\midrule
InstaNAS \cite{cheng2020instanas} & \acc, \lat & Arch+Routing RL Search & - & \makecell[l]{ CIFAR-10) \acc: 95.70\%, \lat: 85ms;\\ CIFAR-100) \acc: 75.80\%, \lat: 89ms, \\ TinyImageNet) \acc: 58.6\%, \lat: 171ms \\ \textit{Device: Intel Core i5-7600 CPU}}
\\
\bottomrule
\end{tabular}%
}
\end{table*}
 
\subsection{NAS with Model Compression}
 
A large body of NAS research integrates model compression—such as pruning, quantization, and architecture scaling—directly into the search loop to jointly optimize accuracy and efficiency. The works are summarized in Table \ref{tab:compression_nas_methods}.
 
\textbf{MSuNAS}~\cite{lu2020nsganetv2} adopts a bi-objective evolutionary strategy over OFA-style supernets (e.g., MobileNetV3~\cite{mobilenetv3}, ProxylessNAS~\cite{cai2019proxylessnas}), optimizing accuracy together with parameters, MACs, or latency. It explores depth, width, kernel size, and input resolution, dynamically selecting surrogate predictors (Radial Basis Function \cite{rbf}, Multi-Layer Perceptron~\cite{mlp}, CARTS~\cite{carts}). 
 
\textbf{CNAS}~\cite{cnas} extends evolutionary search by incorporating technological (memory, compute) and functional (operator) constraints, making it suitable for TinyML and homomorphic-encryption scenarios.
 
A complementary direction constrains differentiable NAS.  
\textbf{RC-DARTS}~\cite{jin2019rc} enforces FLOPs and parameter limits with projection methods, while \textbf{HardCore-NAS}~\cite{hardcorenas} applies differentiable latency constraints using the BC-SFW algorithm~\cite{frankwolfe}. 
\textbf{ZO-DARTS++}~\cite{zodarts++} further integrates parameter constraints using zero-order gradient approximation, avoiding costly backpropagation through architecture parameters.
 
Hardware- and mobile-aware NAS methods form another major group.  
\textbf{FBNet}~\cite{wu2019fbnet} later enabled differentiable latency optimization using LUT-based predictors, training a stochastic supernet to achieve strong accuracy–latency trade-offs.  
\textbf{ProxylessNAS}~\cite{cai2019proxylessnas} removed proxy datasets entirely by training a full over-parameterized supernet and optimizing a differentiable latency model.  
\textbf{OFA}~\cite{ofa} further amortizes search by training a once-for-all network spanning many sub-networks across depth, width, and kernel size, enabling specialized models to be derived in constant time using a pretrained predictor.
 
Model compression can also be embedded explicitly into the search.  
\textbf{DNAS}~\cite{wu2018mixed} performs differentiable bitwidth search at the block level using a supernet with parallel quantized edges, effectively selecting mixed-precision CNNs.  
\textbf{APQ}~\cite{wang2020apq} evolves architectures jointly with pruning and quantization policies via a quantization-aware predictor and LUT-based latency/energy models.

\textbf{HAPFNAS} \cite{hapfnas} extends hardware-aware NAS to federated scenarios, where resource constraints are heterogeneous across clients (per-client FLOPs budgets), using random-forest-based surrogate predictors calibrated on local data distributions, and searching on top of a pretrained supernet. 
 
For microcontroller-class devices, \textbf{MicroNets}~\cite{banbury2021micronets} use differentiable NAS with latency and memory (SRAM/eFlash) constraints, supporting sub-byte quantization. They produce architectures optimized specifically for TinyML workloads.

Lastly, NAS can also compress LLMs, yielding better accuracy-efficiency trade-offs than expert-designed models. \textbf{LoNAS} \cite{munoz-etal-2024-lonas} applies weight-sharing NAS to elastic LoRA adapters \cite{lora}, used as width-pruning masks, attached to frozen Transformer blocks, using NSGA-II to explore the Pareto front between accuracy and computational cost. Its extension, which we name \textbf{LoNAS-2} \cite{compressingllm}, further adds layer-level pruning and a calibrated grid sampling strategy, enabling the removal of entire Transformer blocks. Complementarily, \textbf{Composer} \cite{composer} from Meta FAIR searches over the interleaving pattern and ratio of computational primitives (Attention and MLP) via Bayesian Optimization, discovering novel LLM architectures beyond the fixed 1:1 sequential stacking of standard Transformers. As noted in \cite{lowrankadaptersmeetneural}, the synergy between low-rank representations and NAS is bidirectional: NAS improves adapter quality, while low-rank guidance reduces search cost. Nevertheless, the memory and bandwidth demands of LLMs remain far beyond typical TinyML budgets, making complementary techniques such as quantization \cite{qlora} essential for edge deployment.

\subsection{Neural Architecture Search for Dynamic Neural Networks}
 
\label{subsec:earlyexits}
 
This section reviews NAS approaches targeting Dynamic Neural Networks that jointly design the NN architecture and its adaptive scheme. These works are summarized in Table \ref{tab:dynamic_nas_methods}.
 
An early work is \textbf{EExNAS} \cite{eexnas}, which optimizes an EENN with a single early exit classifier (EEC) placed at a fixed depth, focusing on medical applications.
 
\textbf{HADAS} \cite{hadas} jointly designs the backbone, EECs, and DVFS settings through a two-level genetic search. First, it selects efficient backbones from an OFA supernet \cite{ofa}; then it optimizes the number and placement of early exits, training them with knowledge distillation.
 
 
\textbf{NASEREX} \cite{naserex} learns where and how many exits to attach to a fixed backbone by using evolutionary optimization to improve the accuracy-latency trade-off targeting big image streams.
 
Only a few NAS methods jointly design both backbones and early exits while considering computational constraints. These works rely on genetic search by using NSGA-II \cite{nsga-II} and surrogate predictors to fasten evaluation of models. \textbf{EDANAS} \cite{gambella2023edanas} formulates this as a multi-objective optimization over accuracy and MACs, to select both the architecture and exit parameters. 
\textbf{NACHOS} \cite{nachos} searches for Pareto-optimal EENNs by designing the backbone, EEC positions, thresholds, and EEC architectures under a MACs upper bound. \textbf{AEBNAS} \cite{aebnas} tackles the same constrained problem of NACHOS, but also designs the depth and types of layers in the exit branches in the joint optimization of the NAS.
\textbf{Zniber et al.}~\cite{zniber2025hardwareawareneuralarchitecturesearch} went beyond the optimization through NAS of the MACs of EENNs, proposing a hardware-aware NAS framework for EENNs designed for edge devices, that integrates the effects of quantization and hardware allocation into the design and training process of EENN, ensuring adherence to modern edge accelerators’ deployment constraints. 

While most works focus on depth-wise early exits, \textbf{S2DNAS} \cite{s2dnas} introduces multi-stage structures with exits obtained through channel splitting and feature reuse. \textbf{NASCaps} \cite{nascaps} explores hybrid CapsNet/CNN EENNs, finding Pareto-optimal solutions across accuracy and hardware efficiency using NSGA-II.
 
Finally, \textbf{InstaNAS} \cite{cheng2020instanas} paves the way for NAS targeting Multi-Branch Neural Networks (MBNNs). It jointly optimizes a supernet and a controller that dynamically selects per-sample routing paths—effectively performing dynamic inference. However, it currently supports only static datasets, leaving room for extension to dynamic or streaming scenarios.
 
 
\subsection{NAS for Hardware-Software Co-Design}
 
Hardware-aware NAS increasingly adopts co-design, jointly optimizing neural architectures and accelerator parameters (e.g., PE-array dimensions, dataflows, buffer sizes, tiling schedules). Because these choices strongly influence latency, energy, and throughput, jointly searching network and hardware often yields Pareto-efficient solutions unattainable when either side is fixed. However, the combined search space is large and irregular, motivating the use of supernets, surrogate models, and differentiable relaxations. These works are summarized in Table \ref{tab:codesign_nas_methods}.
 
Early methods like \textbf{Codesign-NAS} performed full training and accelerator simulation for each candidate CNN–hardware pair, but this quickly became impractical \cite{codesignnas}, requiring 96 days for a single GPU. Faster strategies approximate accuracy through supernets, as used in \textbf{NAAS}~\cite{lin2021naas}. Representative methods such as \textbf{EDD}~\cite{li2020edd} and \textbf{DNA}~\cite{zhang2020dna} extend DARTS~\cite{liu_darts_2019} and FBNet~\cite{wu2019fbnet} by embedding accelerator-level choices—parallelism, tiling, buffer sizes, and dataflows—into the optimization loop, enabling efficient end-to-end co-design.
 
RL- and evolutionary-based frameworks explore accelerator templates (FPGA, systolic arrays, ASICs) using analytical or surrogate models for latency and energy. \textbf{AutoDNN}~\cite{hao2019fpga_dnn_codesign} is representative of this direction, while \textbf{NAHAS}~\cite{zhou2021rethinking_codesign} focuses on EdgeTPU by defining a co-search space spanning fused inverted bottlenecks and microarchitectural parameters.
 
More complex deployments include multi-FPGA or heterogeneous systems. \textbf{FNAS}~\cite{jiang2019fpga_aware_nas} searches pipeline partitions across multiple FPGA devices.
 
Co-design also targets emerging device technologies. 
\textbf{NAS4RRAM}~\cite{yuan2021nas4rram} integrates RRAM-specific latency and energy models into the search.
 
Overall, neural–accelerator co-design has evolved into a unified framework where NAS, hardware modeling, and differentiable or surrogate-based optimization cooperate to produce architectures tailored to specific platforms and constraints.
 
\subsection{The weaknesses of these solutions}
\label{subsec:drawback_eff}

While existing hardware-aware NAS solutions have successfully pioneered the design of accurate and lightweight models tailored to specific hardware targets, they fundamentally adopt a static view of the operational environment. In practice, deployed models are subject to two distinct but equally critical forms of degradation that traditional NAS frameworks fail to anticipate.
The first is a lack of robustness to environmental variations, such as the hardware-induced degradation over time. Physical memory devices such as PCM exhibit conductance drift, causing the stored synaptic weights to shift progressively after programming \cite{ielmini2007drift}. This phenomenon alone can drive a model's inference accuracy from 93.75\% down to random guessing (10\%) on CIFAR-10 within 1000 seconds \cite{accuratednnpcm} — a collapse that no amount of search-time optimization can prevent if drift resilience is not explicitly accounted for during architecture design.
The second is distributional degradation over tasks. When a model is deployed in a continual learning setting, the sequential arrival of new tasks induces catastrophic forgetting, causing performance on previously learned tasks to deteriorate sharply \cite{types-incremental}. NAS methods that optimize solely for accuracy on a fixed distribution at search time are structurally blind to this phenomenon: the architectures they produce may be highly accurate and efficient in isolation, yet become fragile or obsolete when confronted with a non-stationary input stream \cite{archcraft}.
Together, these two failure modes reveal a fundamental gap in current hardware-aware NAS: optimizing for accuracy and cost under a static, noise-free assumption is necessary but insufficient for real-world deployment.
 
\section{NAS for robust NN architectures}
\label{sec:robust_nas}

In this Section, we discuss the NAS solutions to find robust NN architectures able to operate in environments with variability.
The outline of this section is the following.
Techniques for robust optimization are introduced in Section \ref{subsec:background_robustness}.
The problem formulation of these NAS solutions is described in Section \ref{subsec:overview_static_dynamic}.
NAS methods for flatness optimization are shown in Section \ref{subsec:flatness}.
NAS methods for adversarial robustness are shown in Section \ref{subsec:adversarial}. NAS procedures for IMC systems are discussed in Section \ref{subsec:imc}. NAS techniques to find NNs robust over multimodalities are explained in Section \ref{subsec:multimodality}. The results of the examined NAS solutions are summarized in Table \ref{tab:robust_nas_methods}.
 
\subsection{Introduction}
\label{subsec:background_robustness}

Robust-aware NAS methods seek architectures that maintain 
high performance despite environmental variations. We 
distinguish five perturbation sources relevant to NAS.

\emph{Flat minima} improve generalization by avoiding sharp 
regions of the loss landscape. Techniques like SAM~\cite{%
sharpnessaware, pittorino22a} explicitly seek flat minima, 
and several NAS methods leverage this principle to stabilize 
the search \cite{GeNAS, a2m}.

\emph{Adversarial robustness} addresses intentional 
perturbations~\cite{szegedy2013intriguing}. Gradient-based 
attacks (FGSM, PGD) find worst-case input directions, while 
adversarial training~\cite{madry2018towards}---minimizing 
loss on adversarially perturbed data---remains the most 
effective defense.

\emph{IMC hardware noise} arises from intrinsic 
non-idealities of emerging non-volatile memories (PCM, RRAM, 
MRAM), including conductance drift, random telegraph noise 
(RTN), and stochastic switching~\cite{ielmini2007drift}. 
Mitigations span hardware-level material 
engineering~\cite{pistolesi2024}, circuit-level 
schemes~\cite{close2010multi}, and algorithmic noise-aware 
training~\cite{ambrogio2019, joshi2020accurate}.

\emph{Resource-shift robustness} is required in 
energy-harvesting scenarios where available power changes 
dynamically~\cite{inas, tinas}. Early-exit networks are 
particularly suited here, as they can adjust the 
accuracy--efficiency trade-off on the fly~\cite{harvnet}. 
A related reliability challenge concerns confidence 
calibration: single-model confidence scores are inherently 
miscalibrated~\cite{squad}, making threshold-based early 
exits fragile even in the absence of adversarial 
perturbations.

\emph{Multimodal robustness} leverages cross-modal 
fusion~\cite{feng2019learning} to handle unreliable 
single-modality inputs. Model merging~\cite{matena2022merging} 
and transformer-based cross-attention~\cite{Li2019VisualBERT} 
offer complementary fusion strategies, while transferability 
across datasets serves as a key robustness proxy in NAS 
benchmarks.

\subsection{Problem formulation}
\label{subsec:overview_static_dynamic}
 
The problem formulation is the following:
 
\begin{align}
\label{eq:adacnas_problem}
    \text{minimize}\quad & \mathcal{G}(F_A(\tilde{x}),F_R(\tilde{x}, \rho, \Delta)) \\
    \text{s.t.}\quad     & \tilde{x} \in \Gamma_{\tilde{x}} \nonumber \\
    & F_A(\tilde{x}) > \bar{A}
    \nonumber \\
                         & F_R(\tilde{x}, \rho, \Delta) > \bar{R} \nonumber
\end{align}

where $\mathcal{G}$ is a multi-objective optimization function, $\tilde{x}$ and $\Gamma_{\tilde{x}}$ represent a candidate NN architecture and the corresponding search space consisting of the architectural parameters, respectively, $F_A(\tilde{x})$ is the accuracy under no perturbations of the environment of $\tilde{x}$ and $\bar{A}$ is the related constraint, and $F_R(\tilde{x},\rho, \Delta)$ is the robustness of the model and $\bar{R}$ is the related constraint, where $\rho$ refers to the intensity of the set of \textbf{environmental perturbations} defined by $\Delta$.
 
Different from the problem formulation of the standard NAS, introduced in Eq. \ref{eq:basenas_problem}, this problem is built on a refined search space, integrates the robustness function, and applies constraints if needed.

\begin{table*}[p]
\caption{Summary of NAS methods for robust model architectures. Objectives marked with an overline ($\overline{\text{Obj}}$) represent hard constraints. Search Cost is in GPU days/hours. Results show the metrics evaluated. The dataset followed by ')' indicates the one where the NAS performed the search, while the datasets between round brackets indicate a dataset to evaluate the transferability.}
\label{tab:robust_nas_methods}
\centering
\resizebox{\textwidth}{!}{%
\begin{tabular}{l l l l l}
\toprule
\textbf{Model} & \textbf{\makecell[l]{Objectives /\\ Constraints}} & \textbf{Method (Robustness)} & \textbf{Search Cost} & \textbf{Results} \\
\midrule
\multicolumn{5}{c}{\textit{\textbf{Flatness-based Generalization}}} \\
\midrule
R-DARTS \cite{Zela2020Understanding} & \acc, \rob(flat) & DARTS+Hessian reg. & $\sim$1 GPU-d & \makecell[tl]{CIFAR-10) \acc: 97.05\%; CIFAR-100) \acc: 81.99\%; \\ SVHN) \acc: 97.83\%; PTB) \acc: 42.41\%} \\
\midrule
GeNAS \cite{GeNAS} & \acc, \rob(flat) & \makecell[tl]{DARTS+Weight noise\\injection} & $\sim$0.4 GPU-d & \makecell[tl]{ImageNet) \acc: 76.10\%, \params: 5.2M, \flops: 0.58G; \\ \acc: 63.38\% (ImageNetV2), \acc: 5.65\% (ImageNetA); \\ \ap: 37.05\% (COCO), \iou: 72.58\% (Cityscapes)} \\
\midrule
NA-DARTS \cite{neighborhood-aware} & \acc, \rob(flat) & DARTS+Neighbors' eval & $\sim$1.1 GPU-d & \makecell[tl]{CIFAR-10) \acc: 97.37\%, \acc: 83.52\% (CIFAR-100); \\ \params: 3.2M; ImageNet) \acc: 74.5\%, \params: 4.8M} \\
\midrule
SDARTS \cite{chen20} & \acc, \rob(flat) & \makecell[tl]{DARTS+Arch Perturbation\\(random/adv)} & $\sim$1.1 GPU-d & \makecell[tl]{CIFAR-10) \acc: 97.52\%, \params: 3.4M, \acc: 75.8\%; \\ PTB) \acc: 43.9\% \params: 23M} \\
\midrule
Self-Distilled \cite{selfdistill} & \acc, \rob(flat) & DARTS+Self Distillation & $\sim$0.37 GPU-d & \makecell[tl]{CIFAR-10) \acc: 97.42\%, \params: 3.3M; \\ ImageNet) \acc: 75\%, \params: 4.7M} \\
\midrule
A$^2$M \cite{a2m} & \acc, \rob(flat) & \makecell[tl]{DARTS with SAM-style\\arch. update} & $\sim$1.3 GPU-d & \makecell[tl]{CIFAR-10) \acc: 97.42\%, \acc: 83.60\% (CIFAR-100); \\ \acc: 58.92\% (ImageNet16-120)} \\
\midrule
\multicolumn{5}{c}{\textit{\textbf{Adversarial Robustness}}} \\
\midrule
ADVRUSH \cite{advrush} & \acc, \rob(flat) & \makecell[tl]{DARTS+Hessian\\approx.} & - & \makecell[tl]{CIFAR-10) \acc: 87.30\%, \rob(PGD-100): 52.80\%; \\ CIFAR-100) \acc: 58.73\%, \rob(PGD-20): 30.15\%; \\ SVHN) \acc: 96.53\%, \rob(PGD-20): 91.14\%; \\ TinyImgNet) \acc: 45.42\%, \rob(PGD-20): 23.58\%} \\
\midrule
RNAS \cite{rnas} & \acc, \rob(adv) & DARTS+noisy samples & $\sim$4.3 GPU-d & \makecell[tl]{CIFAR-10) \acc: 86.30\%, \rob(FGSM): 59.59\%, \\ \rob(PGD-20): 52.65\%} \\
\midrule
RACL \cite{racl} & \acc, \rob(adv) & DARTS+Lipschitz reg. & $\sim$0.5 GPU-d & \makecell[tl]{CIFAR-10) \acc: 84.04\%, \rob(PGD-100): 55.32\%; \\ CIFAR-100) \acc: 57.83\%, \rob(PGD-100): 30.15\%; \\ TinyImgNet) \acc: 48.86\%, \rob(PGD-100): 30.63\%} \\
\midrule
ROBNET \cite{robnet} & \acc, \rob(adv) & OFA+Fine-tuning & \makecell[tl]{OFA +\\ Fine-tuning\textsuperscript{*}} & \makecell[tl]{CIFAR-10) \acc: 82.79\%, \rob(PGD-100): 52.57\%; \\ CIFAR-100) \rob: 23.87\%; SVHN) \rob: 55.59\%; \\ TinyImageNet) Top1: 20.87\%} \\
\midrule
REASON \cite{robustenhancement} & \acc, \rob(adv) & \makecell[tl]{Genetic search+attack-\\independent metric} & - & \makecell[tl]{ImageNet) \acc: 78.91\%, \rob(FGSM): 43.12\%, \\ \rob(PGD-100): 18.27\%} \\
\midrule
WsrNAS \cite{widespectrum} & \acc, \rob(adv) & \makecell[tl]{Gradient search + \\noise estimator} & $\sim$3.6--4 GPU-d & \makecell[tl]{CIFAR-10) \acc: 78.30\%, \rob(FGSM): 67.20\%, \\ \params: 4.5M} \\
\midrule
\multicolumn{5}{c}{\textit{\textbf{In-Memory Computing (IMC) \& Hardware Noise}}} \\
\midrule
CMQ \cite{cmq} & \acc, \rob(hw noise) & \makecell[tl]{Grad. Search on RRAM\\crossbar \& bitwidths} & - & \makecell[tl]{CIFAR-10) \rob: 91.11\%, Resource saving 95.57\%, \\ w prec. 1.4, act. prec. 4 \\ \textit{Device: RRAM crossbar}} \\
\midrule
UAE \cite{uae} & \acc, \rob(hw noise) & \makecell[tl]{RL search (Arch+IMC)\\+ noise injection} & $\sim$255 GPU-h & \makecell[tl]{CIFAR-10) \acc: 80.64\%, \\ \rob(hw noise): 78.39\% \\ \textit{Device: IMC accelerator}} \\
\midrule
Genetic NAS \cite{gaIMC1} & \acc, \rob(hw noise) & \makecell[tl]{Genetic Search on\\Arch+Memristor space} & - & \makecell[tl]{MNIST) \acc: 94.97\%, \rob: 85.90\%, \\ \area: 0.779mm$^2$, Power: 54.52 mW \\ \textit{Device: Memristor IMC}} \\
 
\midrule
\multicolumn{5}{c}{\textit{\textbf{Multi-Modality \& OOD}}} \\
\midrule
NAS-OOD \cite{nasood} & \acc, \rob(modes) & \makecell[tl]{Gradient-based +\\domain generator} & - & \makecell[tl]{OoD-FP) \acc: 98.77\%, \params: 2.33M; \\ NICO) \acc: 85.16\%} \\
\midrule
Model Merging \cite{akiba2025evolutionary} & \acc, \rob(modes) & \makecell[tl]{Genetic Search on\\Param/Dataflow space} & - & \makecell[tl]{GMS8k) \acc: 55.2\% (MGSM-JA), \\ \acc: 66.2\% (JP-LMEH), \params: 10B; \\ JA-VG-VQA) \rouge: 20.4, \rouge: 47.6} \\
\midrule
CM-NAS \cite{cmnas} & \acc, \rob(modes) & \makecell[tl]{Gradient Search on a \\ BN-oriented space} & - & \makecell[l]{SYSU-MM01)  \ap: 60.02 (Single-Shot), \\
\ap: 53.45 (Multi-Shot),  \ap: 78.91 (RegDB V-I), \\ \ap: 77.16 (RegDB I-V); \\
RegDB) \ap: 58.62 (SYSU-MM01 Single-Shot),\\ \ap: 52.07 (SYSU-MM01 Multi-Shot),  \ap: 80.32 (V-I), \\ \ap: 78.31 (I-V)} \\
\midrule
MMNAS \cite{mmnas} & \acc, \rob(modes) & \makecell[tl]{Gradient Search for encoder- \\ decoder space} & - & 
\makecell[l]{
VQA-V2) \acc: 71.35\%, RefCOCO) \acc: 87.40\% (TestA), \\ \acc: 77.70\% (TestB), \acc: 81.00\% (RefCOCO+ TestA), \\ \acc: 65.20\% (RefCOCO+ TestB), 
\acc: 75.70\% (RefCOCOg), \\ Flickr30K) \rec(1): 78.30 (Text retrieval), \\ \rec(1): 60.7 (Image retrieval)}
\\
\midrule
MFAS \cite{PerezRua2019MFAS} & \acc, \rob(modes) & \makecell[tl]{Feature Fusion search\\using SMBO} & \makecell[tl]{$\sim$3.4 GPU-h (AV-M); \\ $\sim$9.2 GPU-h (IMDB); \\ $\sim$25 GPU-d (NTU)} & \makecell[tl]{AV-MNIST) \acc: 88.38\%; \\ MM-IMDB) \fscore: 63.00\%; \\ NTU RGB+D) \acc: 90.04\%} \\
\midrule
BM-NAS \cite{bmnas} & \acc, \rob(modes) & \makecell[tl]{Grad. search on fusion\\\& unimodal selection} & \makecell[tl]{ $\sim$0.98 GPU-h (IMDB); \\ $\sim$38.6 GPU-h (NTU)} & \makecell[tl]{MM-IMDB) \fscore: 62.92\%; \\ NTU) \acc: 90.48\%, \params: 0.98M; \\ EgoGesture) \acc: 94.96\%} \\
\midrule
MANAS \cite{manas} & \acc, \rob(modes) & \makecell[tl]{Search for the best config of \\ logical modules for a context- \\aware supernet} & - & \makecell[tl]{Beauty) \dg(5): 0.26, \hit(5): 0.35; \\ Cellphones) \dg(5): 0.28,  \hit(5): 0.38; \\ Grocery) \dg(5):0.26, \hit(5): 0.36 }  \\
\bottomrule
\multicolumn{5}{l}{\footnotesize \textsuperscript{*} RobNet uses a pre-trained supernet (OFA), reducing search cost during fine-tuning.}
\end{tabular}%
}
\end{table*}
 
\subsection{Flatness-based NAS}
\label{subsec:flatness}
Many recent DARTS variants aim to mitigate the performance collapse or discretization gap, where the supernet’s validation performance poorly predicts that of the final discrete architecture—often due to overfitting and excessive reliance on skip connections. A growing body of work attributes this instability to sharp and irregular optimization landscapes, motivating the search for architectures with greater robustness and flatness.
 
Early evidence visualizes the DARTS loss surface and shows that the optimization can converge to sharp minima, harming generalization~\cite{Shu2020Understanding}.
\textbf{R-DARTS}~\cite{Zela2020Understanding} further connects instability to the Hessian of the validation loss, showing that smaller dominant eigenvalues correlate with more stable and robust search dynamics.
 
Several methods explicitly incorporate smoothness-based regularization into the DARTS objective.
\textbf{GeNAS}~\cite{GeNAS} measures smoothness through accuracy drops induced by weight noise, promoting architectures resilient to parameter perturbations.
\textbf{NA-DARTS}~\cite{neighborhood-aware} extends this idea to the architectural domain by evaluating performance stability under small architectural modifications, thus favoring flatter regions in the architecture space.
 
Other improvements promote flatness indirectly.
\textbf{Self-distillation}~\cite{selfdistill} stabilizes the supernet by transferring knowledge across iterations. Regularization of the architectural parameters via noise injection is used in \textbf{SDARTS-RS} and \textbf{SDARTS-ADV}, both of which enhance robustness during the search~\cite{chen20}.
Finally, $\mathbf{A^2M}$ \cite{a2m} introduces architectural perturbations inspired by sharpness-aware minimization, further improving generalization.
 
\subsection{Adversarial}
\label{subsec:adversarial}

In the field of NAS addressing adversarial robustness, a dedicated 
review has been conducted~\cite{empiricalstudyrobustness} and, more 
recently, a standardized benchmark covering clean and robust accuracy 
for adversarially trained networks from the NAS-Bench-201 space has 
been released~\cite{RobustNAS}, providing a common evaluation 
framework for future work in this direction.
 
In the field of NAS addressing adversarial robustness, \textbf{ADVRUSH}~\cite{advrush}, \textbf{RNAS}~\cite{rnas}, and \textbf{RACL}~\cite{racl}, and  extend the popular NAS method named DARTS~\cite{liu_darts_2019}. 
\textbf{ADVRUSH} penalizes large Hessian eigenvalues to encourage smoother input loss landscapes.
\textbf{RNAS} is designed to optimize the adversarial robustness by implementing a regularization process that computes the similarity between outputs from natural and adversarial data. 
\textbf{RACL} employs a regularization approximating the Lipschitz constant derived from the
architecture parameters.
 Differently, \textbf{ROBNET}~\cite{robnet} and \textbf{REASON}~\cite{robustenhancement} employ a Once-For-All (OFA) supernet~\cite{ofa} to identify NNs that exhibit optimal adversarial robustness. ROBNET achieves this goal by fine-tuning the NNs by adversarial training over a few epochs and evaluating their accuracy when subjected to adversarial attacks. Instead, REASON evaluates the adversarial robustness by using the CLEVER score, a metric that shows a strong generality due to its independence from any attack approach.
 \textbf{WsrNAS}~\cite{widespectrum} proposes a NAS solution to improve robustness over a wide range of adversarial noise strengths by using an adversarial noise estimator.

\subsection{AutoML and NAS for In-Memory Computing}
\label{subsec:imc}

AutoML and HW-NAS methods have proven to be promising for investigating the network model search space and the hardware design space of DNNs deployed on IMC architectures. Notably, the review \cite{reviewNASIMC} illustrates the application of AutoML and HW-NAS to the specific features of IMC hardware and compares existing optimization frameworks. Here, we focus on robustness-aware NAS solutions for IMC hardware, while we will examine NAS accounting also for the hardware efficiency in Section \ref{sec:robust_efficient_nas}. 
 
\textbf{CMQ}~\cite{cmq} is a cross-bar aware mixed-precision quantization scheme looking jointly for the optimal quantization range and weight bitwidths by differentiable architecture search to improve accuracy and resistance to Gaussian device variations in RRAM-based accelerators.
\textbf{UAE}~\cite{uae} searches for a CNN by using Reinforcement Learning and accounting for device variations (thermal/shot noises RTN) and programming errors in IMC architecures. It optimizes accuracy and robustness to these hardware non-idealities.
Another work, which we name \textbf{Genetic NAS}, look for a CNN by using a genetic algorithm and accounting for device variations (Gaussian), conductance
deviation, and device failure in IMC architectures \cite{gaIMC1}. They optimize accuracy and robustness to these hardware non-idealities.
 
\subsection{Multi-Modality}
\label{subsec:multimodality}

\textbf{NAS-OOD}~\cite{nasood} is the first NAS method explicitly designed for domain generalization, where the goal is to discover architectures that can generalize from a set of source domains to an unseen target domain.
In this DARTS-based framework, the architecture search is performed jointly with the optimization of a domain augmentation module that synthesizes out-of-distribution (OOD) samples, simulating novel domain shifts. This enables the search procedure to favor architectures that remain stable when exposed to previously unseen domains.
 
Model merging paired with the evolutionary optimization is shown to be successful in finding LLMs able to perform well in different domains by exploring the weight space, how to mix weights among layers, and the data flow space, how to route tokens through layers \cite{akiba2025evolutionary}. \textbf{CM-NAS}~\cite{cmnas} explores a BN-oriented search space to find the optimal separation scheme for cross-modality matching. In particular, the method works on the VI-ReID datasets, accounting for different lighting, camera settings, and pedestrian populations. The following works go beyond by implementing specific operations for multi-modality in their search space. \textbf{MMNAS}~\cite{mmnas} searches for the best multimodal encoder-decoder backbone, while the specialized heads to tackle multimodal tasks are predefined and fixed by using a gradient-based optimization. Another representative work is \textbf{MFAS}~\cite{PerezRua2019MFAS}, which employs SMBO algorithm \cite{Hutter2011SMBO} to search
multimodal fusion strategies given the unimodal backbones.
But as SMBO is a black-box optimization algorithm, every
update step requires a bunch of DNNs to be trained,
leading to the inefficiency of MFAS. Besides, MFAS only
use concatenation and fully connected (FC) layers for unimodal
feature fusion, and the stack of FC layers would be a
heavy burden for computing. 

Different from previous works, \textbf{BM-NAS}~\cite{bmnas} employs a bi-level NAS to search both unimodal feature selection (which intermediate features from each modality to use, how to route them) and multimodal fusion strategy (how to fuse selected features).
 
In the context of recommender systems, \textbf{MANAS}~\cite{manas} trains a supernet through logical modules to dynamically provide a neural network architecture according to the user context, being the user’s interaction or experience history.

\subsection{The weaknesses of these solutions}
\label{subsec:drawback_rob}

Current NAS methods for robustness suffer from two critical pitfalls that limit their real-world applicability.
The first is hardware overhead. Robustness consistently scales with model size: WideResNet-28-10, the standard backbone in adversarial NAS, carries 36.48M parameters against ResNet-18's 11.17M — a 3.27× overhead — for a modest 5\% gain in robust accuracy \cite{robnet}. Multimodal robustness amplifies this further: MMnas \cite{mmnas} operates at 56–76M parameters, depending on depth, an order of magnitude above TinyML deployment budgets.
The second is temporal blindness. Robustness-focused NAS optimises against static distributional phenomena, but ignores how data evolves over time. This is a structural limitation: robustness and adaptability pull in opposite directions, and maximising one systematically impairs the other. A model designed to resist input perturbations at inference time is, by construction, resistant to the weight updates required to absorb new tasks — the very definition of catastrophic forgetting. ArchCraft \cite{archcraft} confirms this experimentally, showing that state-of-the-art NAS methods, including DARTS-based approaches, fail to produce CL-friendly architectures precisely because they cannot adjust the depth, width, and connectivity properties that govern the stability-plasticity balance \cite{surveynascontinual}.
The result is a class of models that are robust but rigid — well-suited to controlled environments, yet fundamentally unsuited to the non-stationary, resource-constrained settings that define real-world deployment.
 
\section{NAS for Continual Learning}
\label{sec:nas_continual}

\subsection{Introduction}
\label{subsec:background_continual}
 
We now introduce the third pillar of AI, which will be used in some of the examined NAS solutions in this section.
A subset of NAS solutions specifically targets architectures capable of incremental expansion, ensuring sustained performance as new challenges arise over time. At the core of Continual Learning (CL) is the fundamental challenge of balancing stability (the retention of previously acquired knowledge) with plasticity (the capacity to adapt to novel tasks) \cite{types-incremental, continual-survey}. Typically, model efficacy in these settings is measured by the average accuracy across all tasks encountered or by the final performance on the most recent task.

A dedicated survey of the intersection between NAS and continual 
learning has been recently provided~\cite{surveynascontinual}, 
covering expansion-based \cite{adaxpert}, regularization-based \cite{ewc, si-baseline}, and distillation-based 
strategies \cite{lwf}. Our treatment deliberately differs from theirs in two aspects. 
First, we observe that few NAS works optimize continual learning in 
isolation in practice: the vast majority of NAS methods that address 
sequential tasks do so jointly with efficiency constraints, and are 
therefore reviewed in Section~\ref{sec:robust_efficient_nas}. 
Second, we extend the scope beyond continual learning to treat 
efficiency and robustness as equally fundamental pillars, a dimension 
absent from~\cite{surveynascontinual} (see Table~\ref{tab:survey_comparison}). 
This section therefore focuses narrowly on the one representative 
work that targets continual learning in isolation, Continual 
NAS~\cite{neuralarchitecturesearchclassincremental}, using it to 
identify the structural limitations that motivate the joint 
optimization of Section~\ref{sec:robust_efficient_nas}.
 
The complexity of CL is further defined by the nature of the data stream:
 
\begin{itemize}
\item \textbf{Data CL}: The most straightforward scenario, where data arrives sequentially but both the label space and input distribution remain relatively stable~\cite{seal}. The model must retain previously acquired knowledge while absorbing new samples without full retraining.
\item \textbf{Domain CL}: The label space remains fixed across sequential steps, but the underlying input distribution shifts (e.g., the same classification task applied to different image corruptions, sensor modalities, or environments). The challenge is to adapt to distributional shifts without forgetting representations learned under previous domains. Standard benchmarks include Permuted MNIST (P-MNIST) and Rotated MNIST (R-MNIST)~\cite{types-incremental}.
\item \textbf{Task CL}: The model learns a sequence of semantically distinct tasks (e.g., disparate datasets with disjoint label spaces). A Task ID is provided during inference, allowing the model to utilize task-specific modules or output heads to disambiguate between tasks~\cite{types-incremental}.
\item \textbf{Class CL}: The most challenging paradigm, where the model must learn new classes sequentially without the aid of a Task ID at test time. This requires the architecture to effectively distinguish between all observed classes across the entire task history, making catastrophic forgetting particularly damaging~\cite{types-incremental}.
\end{itemize}
 
\subsection{Overview of the NAS Methods}

In this section, we discuss NAS solutions for finding neural network architectures suited to Continual Learning, capable of operating over 
sequential tasks while ensuring a good balance between plasticity and 
stability. Unlike the standard NAS formulation of Eq.~\ref{eq:basenas_problem}, 
where $F_A(x)$ measures accuracy on a fixed data distribution, the 
accuracy objective here becomes the \textbf{Average Incremental Accuracy} 
(AIA), defined as:
\begin{equation}
    F_{AIA}(\tilde{x}, T) = \frac{1}{T}\sum_{t=1}^{T} a_t(\tilde{x})
\end{equation}
where $a_t(\tilde{x})$ is the accuracy of architecture $\tilde{x}$ 
on task $t$ evaluated after learning all tasks up to $T$.
Note that $F_{AIA}(\tilde{x}, T)$ reduces to standard accuracy $F_A(\tilde{x})$ 
when $T=1$, making it a natural generalization that encompasses 
single-task NAS as a special case.
The problem formulation for NAS in the CL setting is therefore:
 
\begin{align}
\label{eq:cl_nas_problem}
    \text{maximize}\quad & F_{AIA}(\tilde{x}, T) \\
    \text{s.t.}\quad     & \tilde{x} \in \Gamma_{\tilde{x}} \nonumber \\
                         & F_{AIA}(\tilde{x}, T) > \bar{A} \nonumber
\end{align}
 
where $\tilde{x}$ and $\Gamma_{\tilde{x}}$ represent a candidate neural 
network architecture and the corresponding search space of architectural 
parameters respectively, and $\bar{A}$ is 
the associated minimum accuracy constraint.
 
Compared to the standard NAS formulation of Eq.~\ref{eq:basenas_problem}, 
this problem differs in three key aspects: the search space is refined to 
account for dynamic model expansion, the accuracy objective incorporates 
temporal evolution through AIA, and the constraint (if present) enforces the minimum acceptable performance across all observed tasks.

Two representative works fall in this category: \textbf{Continual NAS} \cite{neuralarchitecturesearchclassincremental} and \textbf{REC} \cite{rec}, each addressing the accuracy-centric CL problem through structurally distinct approaches.
Continual NAS \cite{neuralarchitecturesearchclassincremental} uses reinforcement learning to search for optimal architecture expansions in terms of average incremental accuracy without constraint in a class-incremental setting. Rather than searching over arbitrary architectures from scratch, at each time step the RL meta-controller decides how much to expand the current architecture in width and depth. Stability is addressed by rehearsing on all past data, while plasticity is achieved through selective model expansion, guided by a heuristic function that prevents unnecessary growth. The experimental results on CIFAR-100 show \aia=56.00\% for 2-Class CL and \aia=58.50\% for 10-Class CL. The search time is $\sim$ 93 GPU hours.
REC \cite{rec} takes a complementary approach, decoupling the forgetting prevention mechanism from the expansion strategy. It employs regularized weight consolidation (RWC) to prevent forgetting without storing past data, combining EWC-style penalties with multi-task L2,1-norm regularization to exploit inter-task correlations. When network capacity is insufficient, an RL-based controller expands the architecture via Net2Wider and Net2Deeper operators, and the expanded model is subsequently compressed back to its original footprint via knowledge distillation after each task. Unlike the expansion-based approach of Continual NAS, REC never permanently increases the architecture size, making it structurally nonexpansive. The experimental results show \aia=59.70\% for 10-Class CL with \params=4M on CIFAR-100 and \aia=95.70\% for 10-Domain CL with \params=0.01M on P-MNIST.

The fact that few works belong to this category reflects a broader shift in the literature: purely accuracy-driven 
Continual NAS proved insufficient in practice, motivating the jointly 
accuracy-efficiency optimization approaches discussed in the next section.

\subsection{The weaknesses of these solutions}
 
Despite their structural differences, both works share the same fundamental limitation: they optimize solely for accuracy, ignoring the hardware constraints that govern real-world deployment. Continual NAS relies on unbounded rehearsal storage, sidestepping strict memory budgets, and its heuristic gatekeeper provides no mechanism to enforce a hard architectural budget or to actively search for resource-efficient solutions. REC enforces a fixed final footprint via post-hoc compression, but neither the expansion search nor the regularization objectives account for latency, energy, or memory during the search itself. Both works also conduct their search on isolated datasets, leaving the resulting architectures potentially fragile under real-world distribution shifts or hardware variability. These shared limitations motivate the jointly accuracy-efficiency optimization approaches discussed in the next section.

\section{NAS for Efficiency, Robustness, and Continual Learning}
\label{sec:robust_efficient_nas}

\subsection{Overview of the NAS methods}

This section reviews NAS approaches that jointly optimize efficiency, robustness,
and accuracy, eventually over continual tasks.
Formally, the NAS addresses the following joint optimization problem:
\begin{align}
\label{eq:hercules_problem}
    \text{minimize }  &\mathcal{G} \left(
    F_A(\tilde{x}),
    F_C^1(\tilde{x}),
    \ldots,
    F_C^N(\tilde{x}),
    F_R(\tilde{x},\rho, \Delta)
    \right) \\
    \text{s. t. }  & \tilde{x} \in \Gamma_{\tilde{x}} \nonumber \\
    & F_A(\tilde{x}) > \bar{A} ,\nonumber \\
    & F_C^i < \bar{C_i} , \quad \forall i = 1, \dots, N \nonumber \\
    & F_R(\tilde{x},\rho, \Delta) > \bar{R} \nonumber
\end{align}
where $\mathcal{G}$ is a multi-objective optimization function, $\tilde{x}$ and
$\Gamma_{\tilde{x}}$ represent a candidate NN architecture enhanced with adaptivity
(i.e., dynamic NN) and the corresponding search space consisting of the architectural
parameters and the hyperparameters of the adaptive scheme, respectively,
$F_A(\tilde{x})$ is the accuracy of the model and $\bar{A}$ is
the related constraint, 
$F_C^i(\widetilde{x})$ is a cost function of $\tilde{x}$ and $\bar{C_i}$ is the
related constraint, $F_R(\tilde{x},\rho, \Delta)$ is the robustness of the model
and $\bar{R}$ is the related constraint, where $\rho$ refers to the intensity of the
set of \textbf{environmental perturbations} defined by $\Delta$.
In the context of continual learning, $F_A(\tilde{x})$ is replaced by
$F_{AIA}(\tilde{x}, T) = \frac{1}{T}\sum_{t=1}^{T} a_t(\tilde{x})$, the Average
Incremental Accuracy over $T$ sequential tasks, which reduces to $F_A(\tilde{x})$
when $T=1$. Hence, $\tilde{x}$ can be seen as the adapted version of a base model
$x$ after model adaptation across continual tasks. \AdaSet, used to build the
$\Gamma_{\tilde{x}}$, and introduced in Section~\ref{sec: efficient_nas}, may contain
the expansion directions across continual tasks, as in SEAL~\cite{seal}.

The works reviewed in this section can be grouped into three broad categories,
mirroring the structure of the preceding sections: (i)~methods that jointly optimize
\emph{efficiency and robustness} to static environmental perturbations
(Sections~\ref{subsec:eff_rob_flatness}--\ref{subsec:eff_rob_multimodal});
(ii)~methods that jointly optimize \emph{efficiency and continual learning}
(Section~\ref{subsec:eff_cl}); and (iii)~methods that target \emph{dynamic
adaptivity at runtime}, where resource availability or environmental conditions
vary during inference (Section~\ref{subsec:dynamic_adaptivity}).
While few works optimize all three pillars simultaneously, this progression
motivates the HERCULES framework introduced in Section~\ref{subsec:hercules_framework}.
The results of all examined NAS solutions are summarized in
Table~\ref{tab:robustefficientnas}.

\begin{table*}[ht!]
\centering
\caption{Summary of NAS methods for robustness, efficiency, and continual learning.
Objectives marked with an overline ($\overline{\text{Obj}}$) represent hard
constraints. Search Cost is in GPU days/hours. Results show key trade-off metrics.}
\label{tab:robustefficientnas}
\resizebox{\textwidth}{!}{%
\begin{tabular}{l l l l l}
\toprule
\textbf{Model} & \textbf{Objectives / Constraints} & \textbf{Method} &
\textbf{Search Cost} & \textbf{Results} \\
\midrule
FlatNAS \cite{flatnas} & \makecell[tl]{\acc, \rob(flat), \\ \params / \paramsconstr} &
  \makecell[tl]{Genetic Search, \\ efficient NNs, \\ weight flatness} & - &
  \makecell[tl]{CIFAR-10) \acc: 90.66\%, Avg-C: 73.42\%; \\ \params: 4.34M, \macs: 173.63M \\
  CIFAR-100) \acc: 72.26\%, Avg-C: 51.29\%; \\ \params: 5.15M, \macs: 338.45M} \\
\midrule
NADAR \cite{nadar} & \makecell[tl]{\acc, \rob(adv), \\ \flops / \flopsconstr} &
  \makecell[tl]{Constrained Gradient \\ Search, dilated NNs} & $\sim$1.8--2.4 GPU-d &
  \makecell[tl]{CIFAR-10) \acc: 86.23\%, \rob(PGD20): 53.43\%; \\ \params: 50.6M, \macs: 7.4G \\
  CIFAR-100) \acc: 62.56\%, \rob(PGD20): 28.40\% \\
  TinyImgNet) \acc: 46.22\%, \rob(PGD20): 21.14\%} \\
\midrule
ANAS \cite{ANAS} & \acc, \rob(adv), \macs &
  \makecell[tl]{Gradient Search, \\ optimal pruning} & $\sim$1 GPU-d &
  \makecell[tl]{CIFAR-10) \acc: 82.42\%, \rob(PGD): 51.53\% \\
  CIFAR-100) \acc: 63.48\%, \rob(PGD): 35.21\% \\
  TinyImgNet) \acc: 41.13\%, \rob(PGD): 17.40\%} \\
\midrule
RoHNAS~\cite{rohnas} & \makecell[tl]{\acc, \rob(adv), \energy, \\ \mem, \lat} &
  \makecell[tl]{Genetic Search for \\ CNN/Capsule networks} & $\sim$2000 GPU-h &
  \makecell[tl]{CIFAR-10) \rob(adv noise): 86.07\%, \energy: 38.63mJ, \\
  \lat: 4.47ms, \mem: 11.85MiB \\
  FMNIST) \rob(adv noise): 93.40\%, \\
  \energy: 61.19mJ, \lat: 6.40ms, \mem: 16.82MiB \\
  \textit{Device: ASIC}} \\
\midrule
NAX \cite{NAX} & \makecell[tl]{\acc, \rob(hw noise), \\ \energy, \lat} &
  \makecell[tl]{MCA Co-design, \\ gradient search} & $\sim$32 GPU-h &
  \makecell[tl]{CIFAR-10) \acc: 93.45\%, \rob: 92.22\%; \\
  \energy: 0.51mJ, \lat: 0.38s$\cdot$mm$^2$ \\
  TinyImgNet) \acc: 56.32\%, \rob: 53.81\%; \\
  \energy: 2.31mJ, \lat: 1.11s$\cdot$mm$^2$ \\ \textit{Device: Memristive crossbar (MCA)}} \\
\midrule
Gibbon~\cite{gibbon} & \makecell[tl]{\acc, \rob(hw noise), \\ \edp, \area} &
  \makecell[tl]{Memristor Co-design \\ by evolutionary search} & $\sim$7 GPU-h &
  \makecell[tl]{CIFAR-10) \rob(hw noise): 88.3\%, \edp: 14.33 ms*mJ, \\
  \area: 186.32 mm$^2$; \textit{Device: Memristor accelerator}} \\
\midrule
NACIM \cite{nacim} & \makecell[tl]{\acc, \rob(hw noise), \\ \energy, \lat, \area} &
  \makecell[tl]{IMC Co-design, \\ RL search} & $\sim$60 GPU-h &
  \makecell[tl]{CIFAR-10, Res.) \rob(hw noise): 73.45\%, \area: 1.97mm$^2$ \\
  CIFAR-10, VGG) \rob(hw noise): 93.12\%, \area: 475mm$^2$ \\
  \textit{Device: IMC accelerator (ReRAM/FeFET/STT-MRAM)}} \\
\midrule
AnalogNAS \cite{benmeziane2023} & \makecell[tl]{\acc, \rob(hw noise) \\ / \paramsconstr} &
  \makecell[tl]{PCM Co-design, \\ evolutionary search} & $\sim$17 GPU-min (+train) &
  \makecell[tl]{CIFAR-10) \rob: 95.00\%, \params: 860k \\ VWW) \rob: 91.00\%, \params: 364k \\
  KWS) \rob: 95.20\%, \params: 456k \\ \textit{Device: PCM analog accelerator}} \\
\midrule
Harmonic-NAS \cite{harmonicnas} & \acc, \rob(modes), \energy, \lat & \makecell[tl]{ bi-level hw-aware search for \\ fusion \& unimodal selection}  & - & \makecell[tl]{AV-MNIST) \rob(modes): 95.11\%, \lat: 7.19ms, \energy: 38.14mJ \\ 
MM-IMDB): \rob(modes): 64.27\%, \lat: 13.05ms, \energy: 140.00mJ \\\textit{Device: Jetson AGX }} \\
\midrule
MOENAS \cite{moenas} & \acc, \rob(modes) / \paramsconstr &
  \makecell[tl]{BO Search, \\ MoE experts} & $\sim$299 GPU-h &
  \makecell[tl]{FACET-S) \acc: 90.33\%, \rob(skin): 93.13\% \\
  FACET-XXS) \acc: 91.72\%, \rob(skin): 88.27\%} \\
\midrule
ERSAM \cite{ersam} & \makecell[tl]{\acc, \rob(noise) \\ / \latencyconstr, \energyconstr, $\overline{TT}$} &
  \makecell[tl]{Pareto supernet, \\ hw constraints} & - &
  \makecell[tl]{LibriSpeech) \rob(modes): 85.7\%, \lat: 47ms, \energy: 38mWh \\
  \textit{Device: Pixel 3 Phone}} \\
\midrule
ENAS-S \cite{nas-stability} & \aia, \rob(flat) &  Genetic Search on CL space & $\sim$ 6 GPU-d & CIFAR-100) \aia: 73.50\% (90+10 Class CL) \params: 3.51M \\
\midrule
SEAL \cite{seal} & \makecell[tl]{\aia, \params, \rob(flat)} &
  \makecell[tl]{Genetic Search for the arch. \\ and its model expansion} & - &
  \makecell[tl]{CIFAR-10) \aia: 95.35\% (5-Data CL), \params: 3.65M \\
  CIFAR-100) \aia: 83.94\% (5-Data CL), \params: 4.09M \\
  ImgNet16) \aia: 52.45\% (5-Data CL), \params: 4.06M} \\
\midrule
CLEAS \cite{cleas} & \aia, \params &
  \makecell[tl]{RL to search for a model \\ with neuron-level expansion} & $\sim$1.44 GPU-h &
  \makecell[tl]{P-MNIST) \aia: 96.30\% (10-Domain CL), Avg \params: $\sim$300k; \\
  R-MNIST) \aia: 96.70\% (10-Domain CL), Avg \params:$\sim$ 300k; \\
  CIFAR-100) \aia: 66.40\% (10-Class CL), \params: $\sim$90k} \\
\midrule
BNS \cite{bns} & \aia, \params &
  \makecell[tl]{RL to search for a model \\ with layer expansion} & - &
  \makecell[l]{MNIST) \aia: 99.87\% (5-Class CL), \aia: 99.26\% \\
  (2-Class CL); CIFAR-10) \aia: 91.40\% (5-Class CL), \\
  \aia: 87.64\% (2-Class CL); \\ CIFAR-100)  \aia: 82.39\% (5-Class CL)} \\
\midrule
LearnToGrow \cite{learning-to-grow} & \aia, \params / \paramsconstr &
  Gradient search on CL space & - &
  \makecell[tl]{P-MNIST) \aia: 98.48\% (10-Domain CL), \params: 2.56M, \\
  CIFAR-100) \aia: 76.21\% (10-Class CL)} \\
\midrule
ArchCraft \cite{archcraft} & \aia / \paramsconstr &
  \makecell[tl]{Evolutionary Search, \\ CL space} & - &
  \makecell[tl]{CIFAR-100) \aia: 82.38\% (5-Class CL), 73.91\% (10-Class CL) \\
  ImgNet100) \aia: 52.02\% (5-Class CL), 45.86\% (10-Class CL)} \\
\midrule
PCL \cite{pcl} & \aia, \params &
  \makecell[tl]{Genetic Search for \\ population-based CL} & - &
  \makecell[tl]{CIFAR-100) \aia: 86.40\% (10-Task CL), \\
  \aia: 78.30\% (10-Class CL), Avg \params: 1.45M, \\
  Avg \flops: 15.9M; TinyImageNet) \aia: 45.60\% (10-Task CL), \\
  \aia: 31.60\% (10-Class CL)} \\
\midrule
DynamicOFA~\cite{dynamicofa} & \acc, \rob(latency) &
  \makecell[tl]{Joint Arch Search and \\ Dynamic Controller training} & - &
  \makecell[tl]{ImageNet) \rob(latency): $\sim$77.5\%, \lat: $\sim$63ms (GPU); \\
  \rob(latency): $\sim$77\%, \lat: $\sim$1.15s (CPU); \textit{Device: GPU, CPU}} \\
\midrule
QoS-NAS~\cite{qosnas} & \acc, \rob(fps) &
  Arch+Routing RL Search & - &
  \makecell[tl]{CIFAR-10) \rob(fps): 83.00\% \fps: 20.50 on avg \\
  across five fps distributions} \\
\midrule
TiNAS \cite{tinas} & \makecell[tl]{\acc, \rob(energy), \\ IMO / \latencyconstr} &
  \makecell[tl]{Evolutionary Search, \\ intermittent exec.} &
  \makecell[tl]{$\sim$1.6 GPU-h (HAR); \\ $\sim$3.1 GPU-h (C10); \\ $\sim$31 GPU-h (MSWC)} &
  \makecell[tl]{HAR) Ti-l \rob: 89.75\%, Ti-t \rob: 91.21\% \\
  C10) Ti-l \rob: 74.66\%, Ti-t \rob: 70.18\% \\
  MSWC) Ti-l \rob: 70.31\%, Ti-t \rob: 64.30\%} \\
\midrule
HarvNet \cite{harvnet} & \makecell[tl]{\acc, \rob(energy) \\ / \macsconstr, \memconstr} &
  \makecell[tl]{Genetic Search, \\ Arch+EE, harvesting} &
  \makecell[tl]{$\sim$47 GPU-d (C10); \\ $\sim$92 GPU-d (Speech);\\ $\sim$3--9 GPU-d (others)} &
  \makecell[tl]{MNIST) \rob: 98.5\%, \mem: 6kB, \macs: 14k \\
  CIFAR-10) \rob: 74.4\%, \mem: 118kB \\
  Chars74K) \rob: 77.3\%, \mem: 21kB \\
  SpeechCmd) \rob: 86.8\%, \mem: 26kB} \\
\bottomrule
\end{tabular}%
}
\end{table*}

\subsection{Joint Efficiency and Robustness Optimization}

The works in this group search for architectures that are simultaneously
lightweight and resilient to \emph{static} environmental perturbations, i.e.,
perturbations whose distribution does not change over the deployment lifetime.
Depending on the nature of the perturbation, these works fall into four
sub-categories: flatness-based generalization, adversarial robustness,
hardware-noise robustness in IMC systems, and multimodal robustness.

\subsubsection{Flatness and OOD Generalization}
\label{subsec:eff_rob_flatness}

\textbf{FlatNAS}~\cite{flatnas} improves OOD generalization by optimizing a
combination of clean accuracy, robustness under weight perturbations, and model
size under a budget on the number of parameters of the model. Unlike NAS-OOD, it
does not require domain labels, targeting general robustness to image corruptions
rather than domain shifts.

\subsubsection{Adversarial Robustness under Efficiency Constraints}
\label{subsec:eff_rob_adv}

\textbf{NADAR}~\cite{nadar} strengthens models under FLOPs constraints by expanding
backbones with robustness-oriented layers while accounting for complexity overhead.
\textbf{ANAS}~\cite{ANAS} searches for per-layer widths of convolutional layers via
pruning masks to enhance adversarial robustness and reduce the model size jointly.
\textbf{RoHNAS}~\cite{rohnas} identifies Pareto-optimal hybrid CapsNet/CNN models
that jointly optimize hardware efficiency (energy, latency, memory), adversarial
robustness, and accuracy using the genetic algorithm NSGA-II. 

\subsubsection{Hardware-Noise Robustness in IMC Systems}
\label{subsec:eff_rob_imc}

Several works address robustness and efficiency specifically on In-Memory Computing
(IMC) hardware, where intrinsic device non-idealities (conductance drift, RTN, and
stochastic switching) degrade inference accuracy after deployment.
\textbf{NAX}~\cite{NAX} jointly explores network
(e.g., kernel size) and hardware parameters (e.g., crossbar size), considering
non-idealities such as wire and interface resistances to optimize energy--delay--area
trade-offs. \textbf{Gibbon}~\cite{gibbon} further co-optimizes architecture and
hardware for accuracy, EDP, latency, and area. \textbf{NACIM}~\cite{nacim} extends
this to device, circuit, and architecture co-search under device-level variations,
incorporating noise models into forward passes and leveraging RL or evolutionary
optimization. \textbf{AnalogNAS}~\cite{benmeziane2023} focuses on PCM accelerators,
using an evolutionary search that models conductance drift directly in the evaluation
loop and enforces a parameter budget.

\subsubsection{Multimodal Robustness}
\label{subsec:eff_rob_multimodal}

\textbf{Harmonic-NAS}~\cite{harmonicnas} jointly searches unimodal backbones and
multimodal fusion networks under hardware constraints. It performs a two-tier
optimization: an evolutionary search for unimodal backbones, followed by
differentiable NAS for fusion modules, with hardware-aware loss terms (energy,
latency) at both stages. To design models robust across different skin tones and
lighting conditions, \textbf{MOENAS}~\cite{moenas} searches for the number of
experts per block in MoE systems to design accurate, fair, and robust edge DNNs.
\textbf{ERSAM}~\cite{ersam} proposes a supernet-based NAS framework to create DNNs
capable of counting concurrent speakers in real time on mobile devices. The primary
goal is to maximize accuracy for social ambiance measurement while strictly adhering
to two hardware constraints (latency and energy) and a data constraint (trainable in
less than 8 hours). The searched DNNs are trained to be robust to background noise
and device domain shifts using a more efficient variant of knowledge distillation.
After supernet training, the NAS provides a Pareto front of admissible solutions
sampled from the supernet.

\subsection{Joint Efficiency and Continual Learning}
\label{subsec:eff_cl}

The works in this group search for architectures that can absorb sequential tasks
while keeping parameter growth under control. Compared to the purely
accuracy-driven Continual NAS of Section~\ref{sec:nas_continual},
all methods here treat model size as a first-class objective or constraint,
making them directly applicable to resource-constrained devices. Most of these NAS frameworks explicitly use dynamic expansion of the architecture upon task arrival.

\textbf{ENAS-S}~\cite{nas-stability} applies a genetic search that jointly
optimizes for accuracy and stability, defined as robustness to perturbations,
providing an early example of linking CL quality to architectural resilience.
\textbf{SEAL}~\cite{seal} jointly searches architectures and
expansion policies for data-incremental learning, avoiding repeated per-task
searches and replay buffers, enabling efficient operation on constrained devices.
It explicitly optimizes the incremental accuracy, number of parameters, and
optionally the flatness of the weight landscape.
\textbf{CLEAS}~\cite{cleas} introduces neuron-level expansion through a
reinforcement learning controller that isolates old weights before expanding the
model, optimizing accuracy at the last task and the number of parameters via
regularization. \textbf{BNS}~\cite{bns} similarly uses RL but searches for
layer-level expansion, reporting results across class- and task-incremental
benchmarks. \textbf{Learn-to-Grow}~\cite{learning-to-grow} performs a gradient
search to decide whether to reuse existing layers, adapt them with lightweight
modules, or grow new layers, enforcing an upper bound on the total parameter count.
\textbf{ArchCraft}~\cite{archcraft} integrates evolutionary algorithms with CL to
find the network with the best incremental accuracy under a parameter budget,
simultaneously reducing width and depth whenever the limit is exceeded.
\textbf{PCL}~\cite{pcl} departs from single-model approaches by searching for a
\emph{population} of task-specific models rather than one unified architecture,
optimizing average population accuracy and parameter count jointly.

\subsection{Dynamic Adaptivity across All Three Pillars}
\label{subsec:dynamic_adaptivity}

The works in this group address a qualitatively different challenge: the
environment itself changes \emph{during inference}, either because the required
quality of service shifts or because the available energy budget fluctuates. These
methods therefore optimize not only the static architecture but also runtime
policies that select or reconfigure the network on the fly. This dynamic dimension
touches all three HERCULES pillars simultaneously---efficiency (adapting compute
to available resources), robustness (maintaining accuracy despite resource
fluctuations), and plasticity (supporting operation under non-stationary conditions)
---and thus represents the closest existing approximation to the full HERCULES ideal.

\subsubsection{Runtime Quality-of-Service and Latency Adaptation}
\label{subsec:dynamic_qos}

\textbf{Quality-of-Service-aware NAS}~\cite{qosnas} extends InstaNAS by designing
a supernet able to adapt its routing to different FPS targets at runtime, reporting
consistent accuracy across five distinct throughput distributions.
\textbf{DynamicOFA}~\cite{dynamicofa} derives a family of sub-networks from an OFA
supernet and uses a runtime manager to select the appropriate sub-network depending
on the required latency, achieving near-$77\%$ ImageNet accuracy across the full
latency range of the family.

\subsubsection{Energy-Harvesting and Intermittent Execution}
\label{subsec:dynamic_harvesting}

NAS techniques also support tiny harvesting devices, where the available power
budget fluctuates unpredictably and execution may be interrupted at any time.
\textbf{TiNAS}~\cite{tinas} explores both architectural and execution parameters
(e.g., tile size, loop order) using evolutionary search to
minimize intermittent latency over variable energy budgets and 
incorporates intermittency management overhead (IMO) directly into the NAS
objective, proposing guidelines to reduce the search space and focus resources on
IMO-sensitive components. \textbf{HarvNet}~\cite{harvnet} addresses
energy-harvesting constraints through (i)~HarvNAS, which designs early-exit
architectures under memory and energy limits, and (ii)~HarvSched, which learns
progressive inference policies conditioned on the runtime harvesting status.

\begin{figure*}[h]
\centering
\includegraphics[width=0.8\textwidth]{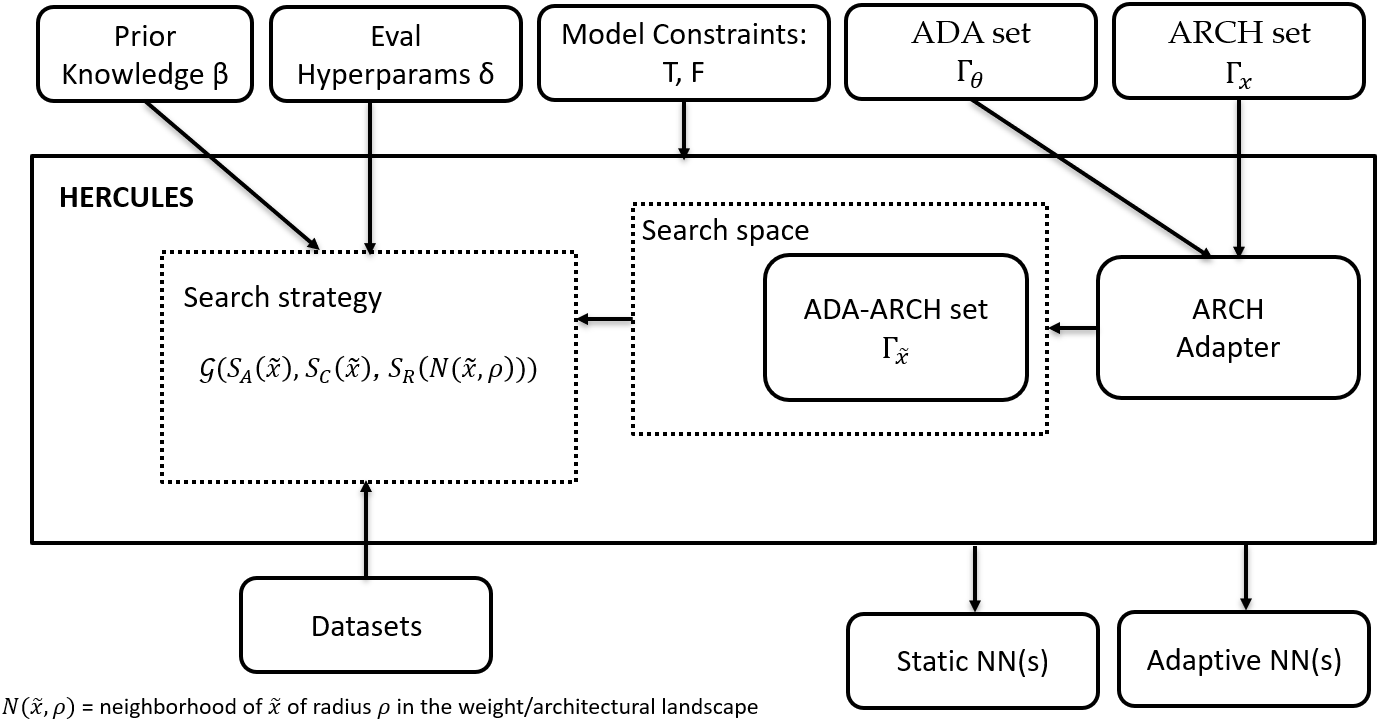}
\caption{Overview of the HERCULES scheme. HERCULES consists of an ARCH Adapter
module, a Search Space, and a Search Strategy.}
\label{fig:framework}
\end{figure*}
 
\section{The HERCULES Framework}
\label{subsec:hercules_framework}
 
Based on the synthesis of existing methodologies, we can finally address our central research question: \textit{Is it possible to automate the design of neural network architectures that maintain performance reliability under hardware noise and data shifts while respecting strict energy budgets and evolving task requirements?}
 
The NAS methods discussed in Section \ref{sec:robust_efficient_nas}, addressing the problem introduced in Eq. \ref{eq:hercules_problem}, pave the way for a new generation of research aimed at discovering neural architectures capable of operating in these demanding real-world edge deployments. Although few existing works optimize all three pillars simultaneously, we argue that the field is rapidly converging toward this necessity. We therefore propose that future AI research adopts a new paradigm that directly answers this question: \textbf{HERCULES (Hardware-Efficient, Robust, and Continual LEarning Search)}.
 
The proposed HERCULES, shown in Figure \ref{fig:framework}, acts as a logical bridge between efficiency, robustness, and continual learning, overcoming the limitations highlighted in Sec. \ref{subsec:drawback_eff} and \ref{subsec:drawback_rob}.
By integrating Adaptivity as the core mechanism, HERCULES
breaks the static nature of traditional NAS. It leverages Dynamic
Neural Networks to activate high-power robustness modules
only when necessary (preserving efficiency) and to reconfigure
architectural paths during sequential tasks (enabling continual
learning). Additionally, the joint optimization of the three objectives ensures that the resulting architectures are not just optimized for a specific hardware at design-time, but are inherently
resilient to the environmental and temporal dynamics of real-world deployment. Lastly, constraints ensure that specific requirements are satisfied and that the search does not converge to a simple balance of the three objectives.
However, this constrained joint optimization of dynamic neural networks requires extra care for the design of a NAS, which we discuss in Section \ref{sec:desiderata}.
 
\subsection{Framework Architecture}
 
An overview of the \textbf{HERCULES} framework is shown in Figure~\ref{fig:framework}. It extends the formalism derived from Hardware-Aware NAS and Robust-aware NAS, introduced in Section \ref{sec: efficient_nas} and \ref{sec:robust_nas}, respectively. It inherits the input of these frameworks but deploys a unified search space and a tri-objective strategy to balance efficiency, robustness, and the temporal evolution of continual learning.
 
The framework takes as input:
\begin{itemize}
  \item A dataset.
  \item A family of neural networks parameterized by an architectural set \OfaSet.
  \item A set of dynamic behavior hyperparameters \AdaSet.
  \item Evaluation hyperparameters \EvalHyper{} (e.g., learning rate and batch size).
  \item Prior information \Prior{}, that can be many empirical measurements, such as the profile of the energy harvesting in TiNAS \cite{tinas} or the conductance drift in IMC \cite{ielmini2007drift}, or the LUTs, which can contain the per-layer latencies measured on a given hardware \cite{proxylessnas}.
  \item A set of constraints:
    \begin{itemize}
      \item \textbf{Technological constraints (\TechConstr)}: Reflect computational and memory limitations.
      \item \textbf{Functional constraints (\FuncConstr)}: Restrict allowable operations, e.g., polynomial operations for encrypted inference.
    \end{itemize}
\end{itemize}
 
The output is a set of \textbf{Pareto-optimal neural networks}, which may be classified as:
\begin{itemize}
  \item \textbf{Dynamic}, if dynamic behavior is incorporated (i.e., \( \AdaSet \neq \emptyset \)). We refer to dynamic architectures as $\tilde{x}$.
  \item \textbf{Static}, if no dynamic behavior is included. We refer to static architectures as $x$.
\end{itemize}
 
 
 
The HERCULES framework is composed of two core modules:
 
\begin{enumerate}
  \item \textbf{Arch Adapter}: Enhances the base network architectures sampled from the ARCH set \OfaSet{} with dynamic capabilities specified by a configuration sampled from the ADA set \AdaSet, and defines the NAS search space ADA-ARCH set \AdaOfaSet. Examples of the ADA set could be the placement and thresholds of early exit classifiers \cite{gambella2023edanas, nachos, aebnas}, or the expansion directions for CL \cite{seal}.
  
  \item \textbf{Search Strategy}: Solves the constrained multi-objective optimization problem using algorithms such as genetic algorithms. The key objectives are:
  \begin{itemize}
    \item \Cost: Resource cost function (e.g., memory usage, number of operations).
    \item $F_R^j(\tilde{x}, \rho_j, \Delta_j)$ for $j = 1, \dots, M$: Robustness functions, one per source of environmental perturbation (e.g., adversarial attacks, hardware noise, distribution shifts). For notational simplicity, the preceding sections used a single $F_R(\tilde{x}, \rho, \Delta)$; HERCULES makes the multi-source nature explicit, since robustness to IMC noise does not imply robustness to adversarial perturbations, and each axis may carry independent constraints.
    \item $F_{AIA}(\tilde{x}, T) = \frac{1}{T}\sum_{t=1}^{T} a_t(\tilde{x})$: Average Incremental Accuracy over $T$ sequential tasks. Note that $F_{AIA}(\tilde{x}, T)$ reduces to standard accuracy $F_A(\tilde{x})$ when $T=1$, so non-CL works are subsumed as a special case without any change of notation. CL-oriented works activate the temporal dimension with $T > 1$.
  \end{itemize}
  The objectives are often not optimized directly but through estimates of surrogate model $S_i(\cdot)$ to speed up the NAS search.
\end{enumerate}
 
 
\section{The twelve labours of HERCULES}
\label{sec:desiderata}
 
We define the HERCULES desiderata as twelve technical challenges. These labours represent the necessary milestones for achieving deployable, lifelong-learning AI systems through the joint optimization of efficiency, robustness, and plasticity. We map these challenges over the three pillars.

\begin{enumerate}
 
    \item \textbf{Hardware-Software Co-design (Efficiency)}: NAS must be integrated with hardware, compilers, and runtimes to ensure actual—not theoretical—performance, moving beyond hardware-agnostic metrics like MACs. This means that the resource cost function \Cost{} should depend on the underlying hardware.
 
    \item \textbf{Scalable Robustness Estimation (Robustness)}: Robustness metrics $F_R(\tilde{x},\rho,\Delta)$ (e.g., adversarial or hardware-noise resilience) must be estimated accurately yet efficiently, avoiding the prohibitive computational overhead (e.g. by performing full-scale attacks for evaluation) during the search phase, possibly by using surrogate models $S_i(\cdot)$.
 
    \item \textbf{Plasticity-Stability Balancing (Continual Learning)}: For the continual learning axis, the framework must explicitly optimize the trade-off between the ability to learn new tasks (plasticity) and the preservation of old knowledge (stability). Hence, we tailor the accuracy $F_A(\tilde{x})$ depending on these two components.
 
    \item \textbf{Deployment Fidelity (Efficiency, Robustness)}: 
    This labour ensures that search-time evaluations faithfully approximate the immediate operational conditions of the target hardware at the moment of deployment. It addresses the immediate gap between theoretical performance and real-world execution. It utilizes Prior information $\beta$, such as empirical measurements of hardware latency and device-specific noise, to model the dynamics of the hardware during the evaluation of $F_{C}(\tilde{x})$ and $F_{R}(\tilde{x},\rho,\Delta)$. This ensures that the selected architecture maintains its robustness and efficiency targets the instant it is transferred from the search environment to the physical accelerator.
    
    \item \textbf{Objective Modeling (All)}: A principled multi-objective formulation that treats efficiency \Cost{}, robustness $F_R(\tilde{x},\rho,\Delta)$, and the performance in continual learning $F_A(\tilde{x})$ as first-class, orthogonal objectives rather than auxiliary regularizers.
    
    \item \textbf{Dynamic Adaptivity (All)}: The search should optimize both static architectural motifs, coming from \OfaSet, and runtime adaptive policies (e.g., early exit), coming from \AdaSet, to handle varying resource availability, environmental conditions, and temporal dynamics. The Arch adapter integrates these two sets, while their optimal synergy is found by the Search Strategy with the aforementioned objectives.
    
    \item \textbf{Expressive Search Spaces (All)}: Search spaces \AdaOfaSet{} must be expressive enough to include architectural blocks that can optimize over all three pillars (i.e., quantization-friendly blocks for increased efficiency). 
    
    \item \textbf{Multi-Objective Search Strategies (All)}: NAS requires optimization algorithms (e.g., evolutionary, reinforcement learning, or differentiable) specifically designed to navigate the non-convex Pareto frontier of the efficiency--robustness--continual learning space in the Search Strategy module.
    
    \item \textbf{Environmental Alignment of the NAS (All)}: NAS should tailor the requirements (and the priorities) according to the environment where the design networks will be deployed. For instance, a noisy environment will require extra care for robustness, and the NAS will select a Search Space \AdaOfaSet{} and a Search Strategy accordingly. Requirements can also be defined by the user-defined constraints (e.g., technological constraints \TechConstr). 
    
    \item \textbf{Benchmark Reproducibility (All)}: Standardized pipelines and benchmarks (e.g., extensions of NAS-Bench \cite{dong2019nasbench201}) are essential to ensure that improvements in the HERCULES axes are comparable across different search algorithms. A unified \textbf{HERCULES-Bench} could combine class-incremental evaluation (e.g., CIFAR-100 with an incremental step of 10 classes), OOD robustness (e.g., CIFAR-10C \cite{benchmarking_ood}), and hardware variability (e.g., the noises on PCM or RRAM simulators). This benchmark should report joint metrics covering the three axes of the HERCULES framework: robustness, plasticity (continual learning), and resource efficiency (MACs and energy).
    
    \item \textbf{Search Efficiency (All)}: The computational cost of the NAS process (Search Cost) itself must be optimized to prevent the search from becoming computationally unfeasible. This is very challenging due to the dimension of the search space and the evaluation of many objectives. Environmental alignment should help by identifying the essential components of the NAS to be activated to perform the search efficiently. 
    
    \item \textbf{Lifecycle Sustainability (All)}: NAS must consider the long-term deployment lifecycle, shifting from a "deploy-and-forget" paradigm to one that accounts for the cumulative costs of architectural updates, re-training for new tasks, and maintenance under aging hardware conditions. This can be seen as the challenge of Deployment Fidelity with a long-term focus and motivates the design of Adaptive NNs over Static NNs, as they can leverage the ADA set $\Gamma_{\theta}$ and the Arch Adapter to provide the necessary plasticity to handle temporal dynamics. By integrating Prior information $\beta$ regarding hardware degradation (e.g., conductance drift in PCM devices), the framework ensures that the searched architectures are inherently sustainable and resilient to environmental and physical evolution over time.
    
\end{enumerate}

\section{Future Directions: Towards Dynamic and Sustainable NAS} \label{sec:future_directions}
 
While the twelve labours represent a joint effort to construct a coherent, unified NAS solution to address the three pillars, further discussion is required to make the search process truly dynamic and sustainable. HERCULES provides deployable dynamic neural networks, yet it remains a static NAS method at its core. Real-world systems, however, encounter evolving data distributions, fluctuating resource budgets, and dynamic user requirements. Future work must therefore integrate adaptability into the NAS process itself, extending the HERCULES principles from static network design to the search mechanism. Crucially, this NAS should grant a runtime version of the \textit{environmental alignment} (Labour 9), able to activate different modules for its search according to the evolving environment conditions in which the NAS is deployed. While the current literature is increasingly becoming mature for the realization of HERCULES, this proposal is a futuristic direction, a further step to unlock the possibility of an auto-evolving HERCULES inside a deployment environment. 
 
A primary direction is Dynamic NAS, where architectures evolve during training in response to environmental changes \cite{dynamicnas}. This work demonstrates how genetic search can redesign the network at each epoch, with the search space adapting through selective perturbative heuristics and weight reuse to ensure training continuity. Future extensions may integrate meta-learned optimizers \cite{metaoptimizers} capable of steering the search under shifting constraints, continual-learning mechanisms to reuse structural knowledge as new tasks arrive \cite{metanas}, and supernets that morph structure or routing paths dynamically—for example, through dynamic subspace sampling or controller distillation \cite{creamcrop, dynamicsubspace}. Lightweight modules such as LoRA adapters \cite{lowrankadaptersmeetneural} also offer promising mechanisms for efficient adaptation during NAS evolution.
 
A second direction involves hardware-integrated learning loops, 
where NAS receives real-time feedback from physical accelerators. 
This would allow architectural decisions to react to drift, noise, 
thermal fluctuations, or energy variations~\cite{ielmini2007drift, joshi2020accurate}, enabling co-design frameworks such as 
crossbar-aware differentiable NAS under realistic IMC 
feedback~\cite{reviewNASIMC, benmeziane2023}. 
This ``in-the-loop'' approach is essential for bridging the gap 
between theoretical robustness---achievable via offline noise 
injection during training~\cite{ambrogio2019}---and actual deployment 
reliability, where conductance drift and device variability evolve 
continuously after programming~\cite{ielmini2007drift, pistolesi2024}. 
Simulation environments such as AIHWKit~\cite{benmeziane2023} 
represent a first step toward this paradigm by providing 
hardware-calibrated noise models, but a full closure of the loop 
on physical hardware, where NAS reacts to measurements from a 
real accelerator rather than a simulator, remains an open challenge.

Another opportunity lies in foundation models for NAS, where pretrained architecture generators learn transferable structural priors to rapidly adapt to new tasks, hardware constraints, or robustness requirements \cite{lapt, llmatic}. Such generators could replace the heavy from-scratch search process with a rapid adaptation phase, significantly lowering the Herculean computational cost of the search.
 
 
\section{Conclusion}
\label{sec:conclusion}
 
This survey has provided a comprehensive overview of the evolving landscape of Neural Architecture Search, moving from traditional accuracy-centric methods to a multi-faceted evaluation of model performance. By reviewing the state-of-the-art across diverse domains, we have demonstrated that while NAS research has traditionally treated efficiency, robustness, and continual learning as separate objectives, these dimensions are fundamentally interconnected. Our analysis of existing literature reveals that modern AI applications—operating under fluctuating computational budgets, non-stationary data distributions, and hardware-induced noise—can no longer rely on disjoint optimization strategies.
 
Through our review of the foundations of NAS, we discussed the mechanisms of efficiency via model compression and hardware--software co-design, examined robustness-oriented NAS across adversarial and environmental perturbations, and explored the architectural requirements for plasticity in sequential task learning. By synthesizing these disparate research threads, we identified a clear gap in current methodologies: the lack of a unified perspective that treats all three pillars as equally fundamental.
 
To address this gap, this survey concludes by proposing HERCULES (Hardware-Efficient, Robust, and Continual LEarning Search). This framework serves as a new paradigm for next-generation NAS, explicitly formalizing the trade-offs between hardware constraints, environmental resilience, and lifelong adaptation. We have outlined the desiderata for HERCULES, highlighting that future progress requires not only more expressive search spaces but also standardized benchmarks—such as the proposed HERCULES-Bench—to evaluate architectures under dynamic, real-world conditions.
 
Ultimately, this survey acts as a roadmap for the field. By shifting the focus toward the HERCULES triad, we provide a conceptual foundation for developing neural architectures that are not only accurate but also resilient, resource-aware, and capable of evolving alongside the demands of safety-critical and resource-limited environments.
 
\bibliographystyle{IEEEtran}
\bibliography{IEEEabrv,
main_no_url.bib,
flatness_no_url.bib,
adversarial_no_url.bib,
ood_no_url.bib,
hardware_no_url.bib,
multimodal_no_url.bib,
earlyexits_no_url.bib,
slimmable.bib,
continual_no_url.bib,
multi_branch_no_url.bib,
dynamicnas_no_url.bib,
imc_no_url.bib,
resourceshift_no_url.bib,
tinyml_no_url.bib
}

@misc{rnas,
	title = {Robust {Neural} {Architecture} {Search}},
	language = {en},
	urldate = {2023-12-29},
	publisher = {arXiv},
	author = {Zhu, Xunyu and Li, Jian and Liu, Yong and Wang, Weiping},
	month = apr,
	year = {2023},
	note = {arXiv:2304.02845 [cs]},
}

@misc{nadar,
	title = {Neural {Architecture} {Dilation} for {Adversarial} {Robustness}},
	language = {en},
	urldate = {2023-12-29},
	publisher = {arXiv},
	author = {Li, Yanxi and Yang, Zhaohui and Wang, Yunhe and Xu, Chang},
	month = aug,
	year = {2021},
	note = {arXiv:2108.06885 [cs]},
}

@misc{robnet,
	title = {When {NAS} {Meets} {Robustness}: {In} {Search} of {Robust} {Architectures} against {Adversarial} {Attacks}},
	shorttitle = {When {NAS} {Meets} {Robustness}},
	language = {en},
	urldate = {2023-12-29},
	publisher = {arXiv},
	author = {Guo, Minghao and Yang, Yuzhe and Xu, Rui and Liu, Ziwei and Lin, Dahua},
	month = mar,
	year = {2020},
	note = {arXiv:1911.10695 [cs, stat]},
}

@misc{racl,
	title = {Adversarially {Robust} {Neural} {Architectures}},
	language = {en},
	urldate = {2023-12-29},
	publisher = {arXiv},
	author = {Dong, Minjing and Li, Yanxi and Wang, Yunhe and Xu, Chang},
	month = feb,
	year = {2023},
	note = {arXiv:2009.00902 [cs]},
}

@misc{empiricalstudyrobustness,
	title = {On {Adversarial} {Robustness}: {A} {Neural} {Architecture} {Search} perspective},
	shorttitle = {On {Adversarial} {Robustness}},
	language = {en},
	urldate = {2024-01-05},
	publisher = {arXiv},
	author = {Devaguptapu, Chaitanya and Agarwal, Devansh and Mittal, Gaurav and Gopalani, Pulkit and Balasubramanian, Vineeth N.},
	month = aug,
	year = {2021},
	note = {arXiv:2007.08428 [cs, stat]},
}

@misc{advrush,
	title = {{AdvRush}: {Searching} for {Adversarially} {Robust} {Neural} {Architectures}},
	shorttitle = {{AdvRush}},
	language = {en},
	urldate = {2023-12-29},
	publisher = {arXiv},
	author = {Mok, Jisoo and Na, Byunggook and Choe, Hyeokjun and Yoon, Sungroh},
	month = aug,
	year = {2021},
	note = {arXiv:2108.01289 [cs]},
}

@Article{robustenhancement,
AUTHOR = {Chen, Haojie and Huang, Hai and Zuo, Xingquan and Zhao, Xinchao},
TITLE = {Robustness Enhancement of Neural Networks via Architecture Search with Multi-Objective Evolutionary Optimization},
JOURNAL = {Mathematics},
VOLUME = {10},
YEAR = {2022},
NUMBER = {15},
ARTICLE-NUMBER = {2724},
ISSN = {2227-7390},
DOI = {10.3390/math10152724}
}

@article{widespectrum, 
title={Neural Architecture Search for Wide Spectrum Adversarial Robustness}, 
volume={37}, 
DOI={10.1609/aaai.v37i1.25118}, 
number={1}, 
journal={Proceedings of the AAAI Conference on Artificial Intelligence}, 
author={Cheng, Zhi and Li, Yanxi and Dong, Minjing and Su, Xiu and You, Shan and Xu, Chang}, 
year={2023}, 
month={Jun.}, 
pages={442-451} 
}

@misc{RobustNAS,
      title={Robust NAS under adversarial training: benchmark, theory, and beyond}, 
      author={Yongtao Wu and Fanghui Liu and Carl-Johann Simon-Gabriel and Grigorios G Chrysos and Volkan Cevher},
      year={2024},
      eprint={2403.13134},
      archivePrefix={arXiv},
      primaryClass={cs.LG},
}

@ARTICLE{ANAS,
  
AUTHOR={Li, Yize  and Zhao, Pu  and Ding, Ruyi  and Zhou, Tong  and Fei, Yunsi  and Xu, Xiaolin  and Lin, Xue },
         
TITLE={Neural architecture search for adversarial robustness via learnable pruning},
        
JOURNAL={Frontiers in High Performance Computing},
        
VOLUME={Volume 2 - 2024},

YEAR={2024},

URL={https://www.frontiersin.org/journals/high-performance-computing/articles/10.3389/fhpcp.2024.1301384},

DOI={10.3389/fhpcp.2024.1301384},

ISSN={2813-7337},
}

@article{szegedy2013intriguing,
  title={Intriguing properties of neural networks},
  author={Szegedy, Christian and Zaremba, Wojciech and Sutskever, Ilya and Bruna, Joan and Erhan, Dumitru and Goodfellow, Ian and Fergus, Rob},
  journal={arXiv preprint arXiv:1312.6199},
  year={2013}
}

@inproceedings{madry2018towards,
  title={Towards deep learning models resistant to adversarial attacks},
  author={Madry, Aleksander and Makelov, Aleksandar and Schmidt, Ludwig and Tsipras, Dimitris and Vladu, Adrian},
  booktitle={Proceedings of the International Conference on Learning Representations (ICLR)},
  year={2018}
}

@INPROCEEDINGS{nas-stability,
  author={Du, Xiaocong and Li, Zheng and Sun, Jingbo and Liu, Frank and Cao, Yu},
  booktitle={2021 International Joint Conference on Neural Networks (IJCNN)}, 
  title={Evolutionary NAS in Light of Model Stability for Accurate Continual Learning}, 
  year={2021},
  volume={},
  number={},
  pages={1-8},
  doi={10.1109/IJCNN52387.2021.9534079}
}

@ARTICLE{cleas,
  author={Gao, Qiang and Luo, Zhipeng and Klabjan, Diego and Zhang, Fengli},
  journal={IEEE Transactions on Neural Networks and Learning Systems}, 
  title={Efficient Architecture Search for Continual Learning}, 
  year={2023},
  volume={34},
  number={11},
  pages={8555-8565},
  doi={10.1109/TNNLS.2022.3151511}}

@inproceedings{learning-to-grow,
  title = 	 {Learn to Grow: A Continual Structure Learning Framework for Overcoming Catastrophic Forgetting},
  author =       {Li, Xilai and Zhou, Yingbo and Wu, Tianfu and Socher, Richard and Xiong, Caiming},
  booktitle = 	 {Proceedings of the 36th International Conference on Machine Learning},
  pages = 	 {3925--3934},
  year = 	 {2019},
  editor = 	 {Chaudhuri, Kamalika and Salakhutdinov, Ruslan},
  volume = 	 {97},
  series = 	 {Proceedings of Machine Learning Research},
  month = 	 {09--15 Jun},
  publisher =    {PMLR}
}

@inproceedings{archcraft, series={IJCAI-2024},
   title={Revisiting Neural Networks for Continual Learning: An Architectural Perspective},
   DOI={10.24963/ijcai.2024/514},
   booktitle={Proceedings of the Thirty-ThirdInternational Joint Conference on Artificial Intelligence},
   publisher={International Joint Conferences on Artificial Intelligence Organization},
   author={Lu, Aojun and Feng, Tao and Yuan, Hangjie and Song, Xiaotian and Sun, Yanan},
   year={2024},
   month=aug, pages={4651–4659},
   collection={IJCAI-2024} }

@misc{pcl,
      title={Position: Continual Learning Benefits from An Evolving Population over An Unified Model}, 
      author={Aojun Lu and Junchao Ke and Chunhui Ding and Jiahao Fan and Yanan Sun},
      year={2025},
      eprint={2502.06210},
      archivePrefix={arXiv},
      primaryClass={cs.LG},
}

@article{types-incremental,
    author = {van de Ven, Gido and Tuytelaars, Tinne and Tolias, Andreas},
    year = {2022},
    month = {12},
    pages = {1-13},
    title = {Three types of incremental learning},
    volume = {4},
    journal = {Nature Machine Intelligence},
    doi = {10.1038/s42256-022-00568-3}
}

@ARTICLE{continual-survey,
  author={Wang, Liyuan and Zhang, Xingxing and Su, Hang and Zhu, Jun},
  journal={IEEE Transactions on Pattern Analysis and Machine Intelligence}, 
  title={A Comprehensive Survey of Continual Learning: Theory, Method and Application}, 
  year={2024},
  volume={46},
  number={8},
  pages={5362-5383},
  doi={10.1109/TPAMI.2024.3367329}
}

@misc{neuralarchitecturesearchclassincremental,
      title={Neural Architecture Search for Class-incremental Learning}, 
      author={Shenyang Huang and Vincent François-Lavet and Guillaume Rabusseau},
      year={2019},
      eprint={1909.06686},
      archivePrefix={arXiv},
      primaryClass={cs.LG},
}

@INPROCEEDINGS{rec,
  author={Zhang, Jie and Zhang, Junting and Ghosh, Shalini and Li, Dawei and Zhu, Jingwen and Zhang, Heming and Wang, Yalin},
  booktitle={2020 IEEE Winter Conference on Applications of Computer Vision (WACV)}, 
  title={Regularize, Expand and Compress: NonExpansive Continual Learning}, 
  year={2020},
  volume={},
  number={},
  pages={843-851},
  doi={10.1109/WACV45572.2020.9093585}}

@inproceedings{bns,
 author = {Qin, Qi and Hu, Wenpeng and Peng, Han and Zhao, Dongyan and Liu, Bing},
 booktitle = {Advances in Neural Information Processing Systems},
 editor = {M. Ranzato and A. Beygelzimer and Y. Dauphin and P.S. Liang and J. Wortman Vaughan},
 pages = {20608--20620},
 publisher = {Curran Associates, Inc.},
 title = {BNS: Building Network Structures Dynamically for Continual Learning},
 volume = {34},
 year = {2021}
}

@misc{seal,
      title={SEAL: Searching Expandable Architectures for Incremental Learning}, 
      author={Matteo Gambella and Manuel Roveri},
      year={2025},
      eprint={2505.10457},
      archivePrefix={arXiv},
      primaryClass={cs.LG},
}

@inproceedings{adaxpert,
  title = 	 {AdaXpert: Adapting Neural Architecture for Growing Data},
  author =       {Niu, Shuaicheng and Wu, Jiaxiang and Xu, Guanghui and Zhang, Yifan and Guo, Yong and Zhao, Peilin and Wang, Peng and Tan, Mingkui},
  booktitle = 	 {Proceedings of the 38th International Conference on Machine Learning},
  pages = 	 {8184--8194},
  year = 	 {2021},
  editor = 	 {Meila, Marina and Zhang, Tong},
  volume = 	 {139},
  series = 	 {Proceedings of Machine Learning Research},
  month = 	 {18--24 Jul},
  publisher =    {PMLR}
}

@article{ewc,
    author = {Kirkpatrick, James and Pascanu, Razvan and Rabinowitz, Neil and Veness, Joel and Desjardins, Guillaume and Rusu, Andrei and Milan, Kieran and Quan, John and Ramalho, Tiago and Grabska-Barwinska, Agnieszka and Hassabis, Demis and Clopath, Claudia and Kumaran, Dharshan and Hadsell, Raia},
    year = {2016},
    month = {12},
    pages = {},
    title = {Overcoming catastrophic forgetting in neural networks},
    volume = {114},
    journal = {Proceedings of the National Academy of Sciences},
    doi = {10.1073/pnas.1611835114}
}

@inproceedings{si-baseline,
  title={Continual learning through synaptic intelligence},
  author={Zenke, Friedemann and Poole, Ben and Ganguli, Surya},
  booktitle={International conference on machine learning},
  pages={3987--3995},
  year={2017},
  organization={PMLR}
}

@article{lwf,
  title={Learning without forgetting},
  author={Li, Zhizhong and Hoiem, Derek},
  journal={IEEE transactions on pattern analysis and machine intelligence},
  volume={40},
  number={12},
  pages={2935--2947},
  year={2017},
  publisher={IEEE}
}

@ARTICLE{surveynascontinual,
  author={Shahawy, Mohamed and Benkhelifa, Elhadj and White, David},
  journal={IEEE Transactions on Neural Networks and Learning Systems}, 
  title={Exploring the Intersection Between Neural Architecture Search and Continual Learning}, 
  year={2025},
  volume={36},
  number={7},
  pages={11776-11792},
  doi={10.1109/TNNLS.2024.3453973}}

@misc{lowrankadaptersmeetneural,
      title={Low-Rank Adapters Meet Neural Architecture Search for LLM Compression}, 
      author={J. Pablo Muñoz and Jinjie Yuan and Nilesh Jain},
      year={2025},
      eprint={2501.16372},
      archivePrefix={arXiv},
      primaryClass={cs.LG},
}

@inproceedings{dynamicnas,
author = {Gerber, Mia and Pillay, Nelishia},
title = {Dynamic Neural Architecture Search for Image Classification},
year = {2024},
isbn = {9798400704956},
publisher = {Association for Computing Machinery},
address = {New York, NY, USA},
doi = {10.1145/3638530.3664117},
booktitle = {Proceedings of the Genetic and Evolutionary Computation Conference Companion},
pages = {1554–1562},
numpages = {9},
location = {Melbourne, VIC, Australia},
series = {GECCO '24 Companion}
}

@ARTICLE{nachos,
  author={Gambella, Matteo and Pomponi, Jary and Scardapane, Simone and Roveri, Manuel},
  journal={IEEE Transactions on Neural Networks and Learning Systems}, 
  title={NACHOS: Neural Architecture Search for Hardware-Constrained Early-Exit Neural Networks}, 
  year={2025},
  volume={},
  number={},
  pages={1-14},
  doi={10.1109/TNNLS.2025.3588558}}

@inproceedings{eexnas,
	address = {Boston, MA, USA},
	title = {{EExNAS}: {Early}-{Exit} {Neural} {Architecture} {Search} {Solutions} for {Low}-{Power} {Wearable} {Devices}},
	isbn = {978-1-66543-922-0},
	shorttitle = {{EExNAS}},
	doi = {10.1109/ISLPED52811.2021.9502503},
	language = {en},
	urldate = {2023-04-17},
	booktitle = {2021 {IEEE}/{ACM} {International} {Symposium} on {Low} {Power} {Electronics} and {Design} ({ISLPED})},
	publisher = {IEEE},
	author = {Odema, Mohanad and Rashid, Nafiul and Faruque, Mohammad Abdullah Al},
	month = jul,
	year = {2021},
	pages = {1--6},
}

@incollection{s2dnas,
	address = {Cham},
	title = {{S2DNAS}: {Transforming} {Static} {CNN} {Model} for {Dynamic} {Inference} via {Neural} {Architecture} {Search}},
	volume = {12347},
	isbn = {978-3-030-58535-8 978-3-030-58536-5},
	shorttitle = {{S2DNAS}},
	language = {en},
	urldate = {2023-04-17},
	booktitle = {Computer {Vision} – {ECCV} 2020},
	publisher = {Springer International Publishing},
	author = {Yuan, Zhihang and Wu, Bingzhe and Sun, Guangyu and Liang, Zheng and Zhao, Shiwan and Bi, Weichen},
	editor = {Vedaldi, Andrea and Bischof, Horst and Brox, Thomas and Frahm, Jan-Michael},
	year = {2020},
	doi = {10.1007/978-3-030-58536-5_11},
	note = {Series Title: Lecture Notes in Computer Science},
	pages = {175--192},
}

@inproceedings{harvnet,
author = {Jeon, Seunghyeok and Choi, Yonghun and Cho, Yeonwoo and Cha, Hojung},
title = {HarvNet: Resource-Optimized Operation of Multi-Exit Deep Neural Networks on Energy Harvesting Devices},
year = {2023},
isbn = {9798400701108},
publisher = {Association for Computing Machinery},
address = {New York, NY, USA},
doi = {10.1145/3581791.3596845},
booktitle = {Proceedings of the 21st Annual International Conference on Mobile Systems, Applications and Services},
pages = {42–55},
numpages = {14},
location = {Helsinki, Finland},
series = {MobiSys '23}
}

@INPROCEEDINGS{hadas,
  author={Bouzidi, Halima and Odema, Mohanad and Ouarnoughi, Hamza and Al Faruque, Mohammad Abdullah and Niar, Smail},
  booktitle={2023 Design, Automation \& Test in Europe Conference \& Exhibition (DATE)}, 
  title={HADAS: Hardware-Aware Dynamic Neural Architecture Search for Edge Performance Scaling}, 
  year={2023},
  volume={},
  number={},
  pages={1-6},
  doi={10.23919/DATE56975.2023.10137095}}

@INPROCEEDINGS{naserex,
  author={Kapoor, Aakash and Soans, Rajath Elias and Dixit, Soham and Ns, Pradeep and Singh, Brijraj and Das, Mayukh},
  booktitle={2023 IEEE International Conference on Big Data (BigData)}, 
  title={NASEREX: Optimizing Early Exits via AutoML for Scalable Efficient Inference in Big Image Streams}, 
  year={2023},
  volume={},
  number={},
  pages={5266-5271},
  doi={10.1109/BigData59044.2023.10386502}}

@article{scardapane_why_2020,
	title = {Why {Should} {We} {Add} {Early} {Exits} to {Neural} {Networks}?},
	volume = {12},
	issn = {1866-9956, 1866-9964},
	doi = {10.1007/s12559-020-09734-4},
	language = {en},
	number = {5},
	urldate = {2023-01-23},
	journal = {Cognitive Computation},
	author = {Scardapane, Simone and Scarpiniti, Michele and Baccarelli, Enzo and Uncini, Aurelio},
	month = sep,
	year = {2020},
	pages = {954--966},
}

@misc{dynamax,
      title={DYNAMAX: Dynamic computing for Transformers and Mamba based architectures}, 
      author={Miguel Nogales and Matteo Gambella and Manuel Roveri},
      year={2025},
      eprint={2504.20922},
      archivePrefix={arXiv},
      primaryClass={cs.CL},
}

@inproceedings{edanas,
  title={{EDANAS}: Adaptive Neural Architecture Search for Early Exit Neural Networks},
  author={Gambella, Matteo and Roveri, Manuel},
  booktitle={2023 International Joint Conference on Neural Networks (IJCNN)},
  pages={1--8},
  year={2023},
  organization={IEEE}
}

@article{jacobs1991moe,
  title={Adaptive mixtures of local experts},
  author={Jacobs, Robert A and Jordan, Michael I and Nowlan, Steven J and Hinton, Geoffrey E},
  journal={Neural Computation},
  volume={3},
  number={1},
  pages={79--87},
  year={1991},
  publisher={MIT Press}
}

@misc{capsnets,
      title={Dynamic Routing Between Capsules}, 
      author={Sara Sabour and Nicholas Frosst and Geoffrey E Hinton},
      year={2017},
      eprint={1710.09829},
      archivePrefix={arXiv},
      primaryClass={cs.CV},
}

@inproceedings{nascaps,
  author    = {Alberto Marchisio and Vojtech Mrazek and Muhammad Shafique},
  title     = {{NASCaps}: A Framework for Neural Architecture Search for Capsule Networks},
  booktitle = {Proceedings of the International Conference on Computer-Aided Design (ICCAD)},
  year      = {2020},
  pages     = {1--9},
  doi       = {10.1145/3400302.3415651}
}

@article{rohnas,
  author    = {Alberto Marchisio and Vojtech Mrazek and Andrea Massa and Beatrice Bussolino and Maurizio Martina and Muhammad Shafique},
  title     = {{RoHNAS}: A Neural Architecture Search Framework With Conjoint Optimization for Adversarial Robustness and Hardware Efficiency of Convolutional and Capsule Networks},
  journal   = {IEEE Access},
  volume    = {10},
  pages     = {109043--109068},
  year      = {2022},
  doi       = {10.1109/ACCESS.2022.3214312}
}

@misc{aebnas,
      title={AEBNAS: Strengthening Exit Branches in Early-Exit Networks through Hardware-Aware Neural Architecture Search}, 
      author={Oscar Robben and Saeed Khalilian and Nirvana Meratnia},
      year={2025},
      eprint={2512.10671},
      archivePrefix={arXiv},
      primaryClass={cs.AI},
      url={https://arxiv.org/abs/2512.10671}, 
}

@misc{zniber2025hardwareawareneuralarchitecturesearch,
      title={Hardware-aware Neural Architecture Search of Early Exiting Networks on Edge Accelerators}, 
      author={Alaa Zniber and Arne Symons and Ouassim Karrakchou and Marian Verhelst and Mounir Ghogho},
      year={2025},
      eprint={2512.04705},
      archivePrefix={arXiv},
      primaryClass={cs.CC},
      url={https://arxiv.org/abs/2512.04705}, 
}

@article{a2m,
	author={Gambella, Matteo and Pittorino, Fabrizio and Roveri, Manuel},
	title={Architecture-Aware Minimization (A$^2$M): How to Find Flat Minima in Neural Architecture Search},
	journal={Machine Learning: Science and Technology},
	year={2025},
}

@misc{neighborhood-aware,
	title = {Neighborhood-{Aware} {Neural} {Architecture} {Search}},
	language = {en},
	urldate = {2023-12-29},
	publisher = {arXiv},
	author = {Wang, Xiaofang and Cao, Shengcao and Li, Mengtian and Kitani, Kris M.},
	month = oct,
	year = {2021},
	note = {arXiv:2105.06369 [cs]},
}

@inproceedings{GeNAS,
  title     = {GeNAS: Neural Architecture Search with Better Generalization},
  author    = {Jeong, Joonhyun and Yu, Joonsang and Park, Geondo and Han, Dongyoon and Yoo, YoungJoon},
  booktitle = {Proceedings of the Thirty-Second International Joint Conference on
               Artificial Intelligence, {IJCAI-23}},
  publisher = {International Joint Conferences on Artificial Intelligence Organization},
  editor    = {Edith Elkind},
  pages     = {911--919},
  year      = {2023},
  month     = {8},
  note      = {Main Track},
  doi       = {10.24963/ijcai.2023/101},
}

@article{selfdistill,
title = {Improving Differentiable Architecture Search via self-distillation},
journal = {Neural Networks},
volume = {167},
pages = {656-667},
year = {2023},
issn = {0893-6080},
doi = {https://doi.org/10.1016/j.neunet.2023.08.062},
author = {Xunyu Zhu and Jian Li and Yong Liu and Weiping Wang},
}

@inproceedings{
Shu2020Understanding,
title={Understanding Architectures Learnt by Cell-based Neural Architecture Search},
author={Yao Shu and Wei Wang and Shaofeng Cai},
booktitle={International Conference on Learning Representations},
year={2020},
}

@inproceedings{
Zela2020Understanding,
title={Understanding and Robustifying Differentiable Architecture Search},
author={Arber Zela and Thomas Elsken and Tonmoy Saikia and Yassine Marrakchi and Thomas Brox and Frank Hutter},
booktitle={International Conference on Learning Representations},
year={2020},
}

@InProceedings{chen20,
  title = 	 {Stabilizing Differentiable Architecture Search via Perturbation-based Regularization},
  author =       {Chen, Xiangning and Hsieh, Cho-Jui},
  booktitle = 	 {Proceedings of the 37th International Conference on Machine Learning},
  pages = 	 {1554--1565},
  year = 	 {2020},
  editor = 	 {III, Hal Daumé and Singh, Aarti},
  volume = 	 {119},
  series = 	 {Proceedings of Machine Learning Research},
  month = 	 {13--18 Jul},
  publisher =    {PMLR},
}

@inproceedings{
sharpnessaware,
title={Sharpness-aware Minimization for Efficiently Improving Generalization},
author={Pierre Foret and Ariel Kleiner and Hossein Mobahi and Behnam Neyshabur},
booktitle={International Conference on Learning Representations},
year={2021}}

@InProceedings{pittorino22a,
  title = 	 {Deep Networks on Toroids: Removing Symmetries Reveals the Structure of Flat Regions in the Landscape Geometry},
  author =       {Pittorino, Fabrizio and Ferraro, Antonio and Perugini, Gabriele and Feinauer, Christoph and Baldassi, Carlo and Zecchina, Riccardo},
  booktitle = 	 {Proceedings of the 39th International Conference on Machine Learning},
  pages = 	 {17759--17781},
  year = 	 {2022},
  editor = 	 {Chaudhuri, Kamalika and Jegelka, Stefanie and Song, Le and Szepesvari, Csaba and Niu, Gang and Sabato, Sivan},
  volume = 	 {162},
  series = 	 {Proceedings of Machine Learning Research},
  month = 	 {17--23 Jul},
  publisher =    {PMLR},
}

@ARTICLE{cmq,
  author={Peng, Jie and Liu, Haijun and Zhao, Zhongjin and Li, Zhiwei and Liu, Sen and Li, Qingjiang},
  journal={IEEE Transactions on Computer-Aided Design of Integrated Circuits and Systems}, 
  title={CMQ: Crossbar-Aware Neural Network Mixed-Precision Quantization via Differentiable Architecture Search}, 
  year={2022},
  volume={41},
  number={11},
  pages={4124-4133},
  doi={10.1109/TCAD.2022.3197495}}

@article{NAS4RRAM,
  title     = {NAS4RRAM: neural network architecture search for inference on RRAM-based accelerators},
  author    = {Yuan, Zhihang and Liu, Jingze and Li, Xingchen and Yan, Longhao and Chen, Haoxiang and Wu, Bingzhe and Yang, Yuchao and Sun, Guangyu},
  journal   = {Science China Information Sciences},
  volume    = {64},
  number    = {6},
  pages     = {160407},
  year      = {2021},
  publisher = {Science China Press and Springer-Verlag GmbH Germany, part of Springer Nature},
  doi       = {10.1007/s11432-020-3245-7},
}

@INPROCEEDINGS{uae,
  author={Yan, Zheyu and Juan, Da-Cheng and Hu, Xiaobo Sharon and Shi, Yiyu},
  booktitle={2021 26th Asia and South Pacific Design Automation Conference (ASP-DAC)}, 
  title={Uncertainty Modeling of Emerging Device based Computing-in-Memory Neural Accelerators with Application to Neural Architecture Search}, 
  year={2021},
  volume={},
  number={},
  pages={859-864},
  doi={}}

@INPROCEEDINGS{gaIMC1,
  author={Krestinskaya, O. and Salama, K. and James, A. P.},
  booktitle={2020 IEEE International Symposium on Circuits and Systems (ISCAS)}, 
  title={Towards Hardware Optimal Neural Network Selection with Multi-Objective Genetic Search}, 
  year={2020},
  volume={},
  number={},
  pages={1-5},
  doi={10.1109/ISCAS45731.2020.9180514}}

@article{NAX,
  author       = {Shubham Negi and
                  Indranil Chakraborty and
                  Aayush Ankit and
                  Kaushik Roy},
  title        = {{NAX:} Co-Designing Neural Network and Hardware Architecture for Memristive
                  Xbar based Computing Systems},
  journal      = {CoRR},
  volume       = {abs/2106.12125},
  year         = {2021},
  eprinttype    = {arXiv},
  eprint       = {2106.12125},
  bibsource    = {dblp computer science bibliography, https://dblp.org}
}

@INPROCEEDINGS{gibbon,
  author={Sun, Hanbo and Wang, Chenyu and Zhu, Zhenhua and Ning, Xuefei and Dai, Guohao and Yang, Huazhong and Wang, Yu},
  booktitle={2022 Design, Automation \& Test in Europe Conference \& Exhibition (DATE)}, 
  title={Gibbon: Efficient Co-Exploration of NN Model and Processing-In-Memory Architecture}, 
  year={2022},
  volume={},
  number={},
  pages={867-872},
  doi={10.23919/DATE54114.2022.9774605}}

@ARTICLE{nacim,
  author={Jiang, Weiwen and Lou, Qiuwen and Yan, Zheyu and Yang, Lei and Hu, Jingtong and Hu, Xiaobo Sharon and Shi, Yiyu},
  journal={IEEE Transactions on Computers}, 
  title={Device-Circuit-Architecture Co-Exploration for Computing-in-Memory Neural Accelerators}, 
  year={2021},
  volume={70},
  number={4},
  pages={595-605},
  doi={10.1109/TC.2020.2991575}}

@inproceedings{benmeziane2023,
  author    = {Benmeziane, H. and Lammie, C. and Boybat, I. and Rasch, M. and Le Gallo, M. and Tsai, H. and Muralidhar, R. and Niar, S. and Hamza, O. and Narayanan, V. and Sebastian, A. and El Maghraoui, K.},
  title     = {{AnalogNAS}: A Neural Network Design Framework for Accurate Inference With Analog In-Memory Computing},
  booktitle = {2023 {IEEE} International Conference on Edge Computing and Communications ({EDGE})},
  year      = {2023},
  pages     = {233--244},
  doi       = {10.1109/EDGE60047.2023.00041}
}

@article{jin2019rc,
  title={Rc-darts: Resource constrained differentiable architecture search},
  author={Jin, Xiaojie and Wang, Jiang and Slocum, Joshua and Yang, Ming-Hsuan and Dai, Shengyang and Yan, Shuicheng and Feng, Jiashi},
  journal={arXiv preprint arXiv:1912.12814},
  year={2019}
}

@inproceedings{hardcorenas,
  title={HardCoRe-NAS: Hard Constrained diffeRentiable Neural Architecture Search},
  author={Niv Nayman and Yonathan Aflalo and Asaf Noy and Lihi Zelnik-Manor},
  booktitle={ICML},
  year={2021}
}

@inproceedings{lu2020nsganetv2,
  title={Nsganetv2: Evolutionary multi-objective surrogate-assisted neural architecture search},
  author={Lu, Zhichao and Deb, Kalyanmoy and Goodman, Erik and Banzhaf, Wolfgang and Boddeti, Vishnu Naresh},
  booktitle={European Conference on Computer Vision},
  pages={35--51},
  year={2020},
  organization={Springer}
}

@misc{proxylessnas,
      title={ProxylessNAS: Direct Neural Architecture Search on Target Task and Hardware}, 
      author={Han Cai and Ligeng Zhu and Song Han},
      year={2019},
      eprint={1812.00332},
      archivePrefix={arXiv},
      primaryClass={cs.LG},
}

@inproceedings{hao2019fpga_dnn_codesign,
  author    = {Hao, Cong and Zhang, Xinheng and Li, Yun and Huang, Shuo and Xiong, Jinjun and Rupnow, Kyle and Hwu, Wen{-}Mei and Chen, Deming},
  title     = {{FPGA/DNN Co-Design}: An Efficient Design Methodology for IoT Intelligence on the Edge},
  booktitle = {Proceedings of the 56th ACM/IEEE Design Automation Conference (DAC)},
  year      = {2019},
  pages     = {1--6},
  month     = jun,
  publisher = {ACM},
  doi       = {10.1145/3316781.3317768}
}

@article{zhang2020dna,
  author       = {Zhang, Yiyu and Fu, Yixing and Jiang, Wenxuan and Li, Chang and You, Haoran and Li, Meng and Chandra, Vikas and Lin, Yujun},
  title        = {DNA: Differentiable Network-Accelerator Co-Search},
  journal      = {arXiv preprint arXiv:2010.14778},
  year         = {2020}
}

@inproceedings{li2020edd,
  author    = {Li, Yun and Hao, Cong and Zhang, Xinheng and Liu, Xuechao and Chen, Yiyu and Xiong, Jinjun and Hwu, Wen{-}Mei and Chen, Deming},
  title     = {{EDD}: Efficient Differentiable {DNN} Architecture and Implementation Co-Search for Embedded {AI} Solutions},
  booktitle = {Proceedings of the 57th ACM/EDAC/IEEE Design Automation Conference (DAC)},
  year      = {2020},
  pages     = {1--6},
  month     = jul,
  publisher = {IEEE}
}

@inproceedings{lin2021naas,
  author    = {Lin, Yujun and Yang, Mingyu and Han, Song},
  title     = {{NAAS}: Neural Accelerator Architecture Search},
  booktitle = {Proceedings of the 58th ACM/ESDA/IEEE Design Automation Conference (DAC)},
  year      = {2021},
  pages     = {1--7},
  publisher = {IEEE}
}

@article{zhou2021rethinking_codesign,
  author       = {Zhou, Yuhui and Dong, Xiaoliang and Akin, Berk and Tan, Mingxing and Peng, Dian and Meng, Tao and Yazdanbakhsh, Amir and Huang, Da and Narayanaswami, Ravi and Laudon, James},
  title        = {Rethinking Co-Design of Neural Architectures and Hardware Accelerators},
  journal      = {arXiv preprint arXiv:2102.08619},
  year         = {2021}
}

@inproceedings{jiang2019fpga_aware_nas,
  author    = {Jiang, Wenxuan and Zhang, Xinheng and Sha, Edwin Hsing{-}Mean and Yang, Liang and Zhuge, Qingfeng and Shi, Yiyu and Hu, Jian},
  title     = {Accuracy vs. Efficiency: Achieving Both Through {FPGA}-Implementation-Aware Neural Architecture Search},
  booktitle = {Proceedings of the 56th ACM/IEEE Design Automation Conference (DAC)},
  year      = {2019},
  pages     = {1--6},
  month     = dec,
  publisher = {IEEE}
}

@article{yuan2021nas4rram,
  author    = {Yuan, Zhiyuan and Liu, Jie and Li, Xuehai and Yan, Liqiang and Chen, Hu and Wu, Bo and Yang, Yiran and Sun, Guangyu},
  title     = {{NAS4RRAM}: Neural Network Architecture Search for Inference on {RRAM}-Based Accelerators},
  journal   = {Science China Information Sciences},
  volume    = {64},
  number    = {6},
  year      = {2021},
  articleno = {160407}
}

@article{wu2018mixed,
  title={Mixed Precision Quantization of Convnets via Differentiable Neural Architecture Search},
  author={Wu, Bichen and Wang, Yanghan and Zhang, Peizhao and Tian, Yuandong and Vajda, Peter and Keutzer, Kurt},
  journal={arXiv preprint arXiv:1812.00090},
  year={2018}
}

@inproceedings{wu2019fbnet,
  title={{FBNet}: Hardware-Aware Efficient ConvNet Design via Differentiable Neural Architecture Search},
  author={Wu, Bichen and Keutzer, Kurt and Dai, Xiaoliang and Zhang, Peizhao and Wang, Yanghan and Sun, Fei and Wu, Yiming and Tian, Yuandong and Vajda, Peter and Jia, Yangqing},
  booktitle={Proceedings of the IEEE/CVF Conference on Computer Vision and Pattern Recognition (CVPR)},
  pages={10734--10742},
  year={2019},
  month={Jun.}
}

@inproceedings{wang2020apq,
  title={{APQ}: Joint Search for Network Architecture, Pruning and Quantization Policy},
  author={Wang, Tong and Wang, Kai and Cai, Han and Lin, Ji and Liu, Zhijian and Wang, Haotian and Lin, Yujun and Han, Song},
  booktitle={Proceedings of the IEEE/CVF Conference on Computer Vision and Pattern Recognition (CVPR)},
  pages={2075--2084},
  year={2020},
  month={Jun.}
}

@inproceedings{banbury2021micronets,
  title={{MicroNets}: Neural Network Architectures for Deploying TinyML Applications on Commodity Microcontrollers},
  author={Banbury, Colby and Zhou, Ching and Fedorov, Igor and Navarro, Rumen M. and Thakker, Utkarsh and Gope, Debidatta and Reddi, Vijay Janapa and Mattina, Matthew and Whatmough, Paul N.},
  booktitle={Proceedings of Machine Learning and Systems (MLSys)},
  pages={1--16},
  year={2021}
}

@article{ielmini2007drift,
  author    = {Ielmini, Daniele and Lacaita, Andrea L. and Mantegazza, D.},
  title     = {Recovery and drift dynamics of resistance and threshold voltages in phase-change memories},
  journal   = {{IEEE} Transactions on Electron Devices},
  year      = {2007},
  volume    = {54},
  number    = {2},
  pages     = {308--315},
  doi       = {10.1109/TED.2006.888752}
}

@inproceedings{ambrogio2019,
  author    = {Ambrogio, S. and Gallot, M. and Spoon, K. and Tsai, H. and Mackin, C. and Wesson, M. and Kariyappa, S. and Narayanan, P. and Liu, C.-C. and Kumar, A. and Chen, A. and Burr, G. W.},
  title     = {Reducing the impact of phase-change memory conductance drift on the inference of large-scale hardware neural networks},
  booktitle = {2019 {IEEE} International Electron Devices Meeting ({IEDM})},
  year      = {2019},
  pages     = {6.1.1--6.1.4},
  doi       = {10.1109/IEDM19573.2019.8993565}
}

@inproceedings{pistolesi2024,
  author    = {Pistolesi, L. and Glukhov, A. and de Gracia Herranz, A. and Lopez-Vallejo, M. and Carissimi, M. and Pasotti, M. and Rolandi, P. and Redaelli, A. and Mart{\'\i}n, I. M. and Bianchi, S. and Bonfanti, A. and Ielmini, D.},
  title     = {Drift compensation in multilevel {PCM} for in-memory computing accelerators},
  booktitle = {2024 {IEEE} International Reliability Physics Symposium ({IRPS})},
  year      = {2024},
}

@inproceedings{close2010multi,
  author    = {Close, G. F. and Breitwisch, M. J. and Lam, C. H. and Hagleitner, C. and Nirschl, T. and Lung, H. L. and Zhu, Y.},
  title     = {A 512Mb multi-level phase change memory with 1.2V, 100ns read and 5V, 1us program},
  booktitle = {2010 {IEEE} International Solid-State Circuits Conference Digest of Technical Papers ({ISSCC})},
  year      = {2010},
  pages     = {208--209},
  doi       = {10.1109/ISSCC.2010.5433920}
}

@article{joshi2020accurate,
  author    = {Joshi, Vinay and Le Gallo, Manuel and Haefeli, Simon and Boybat, Irem and Nandakumar, S. R. and Piveteau, Charles and Dazzi, Martino and Rajendran, Bipin and Sebastian, Abu and Eleftheriou, Evangelos},
  title     = {Accurate deep neural network inference using computational phase-change memory},
  journal   = {Nature Communications},
  year      = {2020},
  volume    = {11},
  number    = {1},
  pages     = {2425},
  doi       = {10.1038/s41467-020-16108-9}
}

@article{accuratednnpcm,
  title={Accurate deep neural network inference using computational phase-change memory},
  author={Joshi, Vinay and Le Gallo, Manuel and Haefeli, Simon and Boybat, Irem and Nandakumar, S. R. and Piveteau, Christophe and Dazzi, Martino and Rajendran, Bipin and Sebastian, Abu and Eleftheriou, Evangelos},
  journal={Nature Communications},
  volume={11},
  number={1},
  pages={2473},
  year={2020},
  publisher={Nature Publishing Group}
}

@inproceedings{codesignnas,
author = {Abdelfattah, Mohamed S. and Dudziak, Lukasz and Chau, Thomas and Lee, Royson and Kim, Hyeji and Lane, Nicholas D.},
title = {Codesign-NAS: Automatic FPGA/CNN Codesign Using Neural Architecture Search},
year = {2020},
isbn = {9781450370998},
publisher = {Association for Computing Machinery},
address = {New York, NY, USA},
url = {https://doi.org/10.1145/3373087.3375334},
doi = {10.1145/3373087.3375334},
booktitle = {Proceedings of the 2020 ACM/SIGDA International Symposium on Field-Programmable Gate Arrays},
pages = {315},
numpages = {1},
location = {Seaside, CA, USA},
series = {FPGA '20}
}

@article{nasframework,
title = {Neural architecture search: A contemporary literature review for computer vision applications},
journal = {Pattern Recognition},
volume = {147},
pages = {110052},
year = {2024},
issn = {0031-3203},
doi = {https://doi.org/10.1016/j.patcog.2023.110052},
author = {Matt Poyser and Toby P. Breckon},
}

@misc{ofa,
	title = {Once-for-{All}: {Train} {One} {Network} and {Specialize} it for {Efficient} {Deployment}},
	shorttitle = {Once-for-{All}},
	language = {en},
	urldate = {2023-01-25},
	publisher = {arXiv},
	author = {Cai, Han and Gan, Chuang and Wang, Tianzhe and Zhang, Zhekai and Han, Song},
	month = apr,
	year = {2020},
	note = {arXiv:1908.09791 [cs, stat]},
}

@article{nsga-II,
	title = {A fast and elitist multiobjective genetic algorithm: {NSGA}-{II}},
	volume = {6},
	issn = {1089778X},
	shorttitle = {A fast and elitist multiobjective genetic algorithm},
	doi = {10.1109/4235.996017},
	language = {en},
	number = {2},
	urldate = {2023-01-28},
	journal = {IEEE Transactions on Evolutionary Computation},
	author = {Deb, K. and Pratap, A. and Agarwal, S. and Meyarivan, T.},
	month = apr,
	year = {2002},
	pages = {182--197},
}

@article{elsken2019nas,
  author    = {Thomas Elsken and Jan Hendrik Metzen and Frank Hutter},
  title     = {Neural Architecture Search: A Survey},
  journal   = {Journal of Machine Learning Research},
  volume    = {20},
  pages     = {55:1--55:21},
  year      = {2019},
}

@article{ren2021nas,
  author    = {Pengzhen Ren and Yun Xiao and Xiaojun Chang and Po{-}Yao Huang and Zhihui Li and Xiaojiang Chen and Xin Wang and Zhangjun He},
  title     = {A Comprehensive Survey of Neural Architecture Search: Challenges and Solutions},
  journal   = {ACM Computing Surveys},
  volume    = {54},
  number    = {4},
  pages     = {1--34},
  year      = {2021},
  doi       = {10.1145/3447585}
}

@article{sekanina2021nas,
  author    = {Luk{\'{a}}s Sekanina},
  title     = {Neural Architecture Search and Hardware Accelerator Co-Search: A Survey},
  journal   = {IEEE Access},
  volume    = {9},
  pages     = {151337--151362},
  year      = {2021},
  doi       = {10.1109/ACCESS.2021.3126056}
}

@article{chitty2022nas,
  author    = {Krishna Teja Chitty{-}Venkata and Arun K. Somani},
  title     = {Neural Architecture Search Survey: A Hardware Perspective},
  journal   = {ACM Computing Surveys},
  volume    = {55},
  number    = {8},
  pages     = {1--36},
  year      = {2022},
  doi       = {10.1145/3524500}
}

@inproceedings{benmeziane2021ijcai,
  author    = {Houda Benmeziane and Ruxandra Tudoran and Youn{\`{e}}s Bennani and Gabriele Cabodi and Liliana Pasquale},
  title     = {Hardware-Aware Neural Architecture Search: Survey and Taxonomy},
  booktitle = {Proceedings of the Thirtieth International Joint Conference on Artificial Intelligence},
  pages     = {4322--4329},
  year      = {2021},
  publisher = {International Joint Conferences on Artificial Intelligence Organization},
  doi       = {10.24963/ijcai.2021/592}
}

@article{reviewNASIMC,
  author       = {Krestinskaya, Olga and Fouda, Mo El-Mahdi and Benmeziane, Hassen and James, Alex Pappachen},
  title        = {Neural architecture search for in-memory computing-based deep learning accelerators},
  journal      = {Nature Reviews Electrical Engineering},
  volume       = {1},
  pages        = {374--390},
  year         = {2024},
}

@article{mobilenetv3,
  title={Searching for MobileNetV3},
  author={Andrew G. Howard and Mark Sandler and Grace Chu and Liang-Chieh Chen and Bo Chen and Mingxing Tan and Weijun Wang and Yukun Zhu and Ruoming Pang and Vijay Vasudevan and Quoc V. Le and Hartwig Adam},
  journal={2019 IEEE/CVF International Conference on Computer Vision (ICCV)},
  year={2019},
  pages={1314-1324}
}

@misc{diffusionNAS,
      title={DiffusionNAG: Predictor-guided Neural Architecture Generation with Diffusion Models}, 
      author={Sohyun An and Hayeon Lee and Jaehyeong Jo and Seanie Lee and Sung Ju Hwang},
      year={2024},
      eprint={2305.16943},
      archivePrefix={arXiv},
      primaryClass={cs.LG},
}

@inproceedings{nsganetv2,
  title={Nsganetv2: Evolutionary multi-objective surrogate-assisted neural architecture search},
  author={Lu, Zhichao and Deb, Kalyanmoy and Goodman, Erik and Banzhaf, Wolfgang and Boddeti, Vishnu Naresh},
  booktitle={European Conference on Computer Vision},
  pages={35--51},
  year={2020},
  organization={Springer}
}

@ARTICLE{zeroshot,
  author={Li, Guihong and Hoang, Duc and Bhardwaj, Kartikeya and Lin, Ming and Wang, Zhangyang and Marculescu, Radu},
  journal={IEEE Transactions on Pattern Analysis and Machine Intelligence}, 
  title={Zero-Shot Neural Architecture Search: Challenges, Solutions, and Opportunities}, 
  year={2024},
  volume={46},
  number={12},
  pages={7618-7635},
  doi={10.1109/TPAMI.2024.3395423}}

@inproceedings{
liu_darts_2019,
title={{DARTS}: Differentiable Architecture Search},
author={Hanxiao Liu and Karen Simonyan and Yiming Yang},
booktitle={International Conference on Learning Representations},
year={2019},
}

@misc{rlsurvey,
      title={Reinforcement Learning: A Survey}, 
      author={L. P. Kaelbling and M. L. Littman and A. W. Moore},
      year={1996},
      eprint={cs/9605103},
      archivePrefix={arXiv},
      primaryClass={cs.AI}
}

@inproceedings{zodarts+,
  title     = {An Efficient Neural Architecture Search Model for Medical Image Classification},
  author    = {Lunchen Xie and Eugenio Lomurno and Matteo Gambella and Danilo Ardagna and Manuel Roveri and Matteo Matteucci and Qingjiang Shi},
  booktitle = {Proceedings of the European Symposium on Artificial Neural Networks, Computational Intelligence and Machine Learning (ESANN)},
  year      = {2024},
  pages     = {661--666},
  publisher = {i6doc.com publ.},
  address   = {Bruges, Belgium and online event},
  isbn      = {978-2-87587-090-2},
  note      = {ZO-DARTS+: a differentiable NAS algorithm for efficient medical image classification}
}

@misc{zodarts++,
      title={ZO-DARTS++: An Efficient and Size-Variable Zeroth-Order Neural Architecture Search Algorithm}, 
      author={Lunchen Xie and Eugenio Lomurno and Matteo Gambella and Danilo Ardagna and Manual Roveri and Matteo Matteucci and Qingjiang Shi},
      year={2025},
      eprint={2503.06092},
      archivePrefix={arXiv},
      primaryClass={cs.CV},
}

@misc{zeinaty2025,
      title={Can LLMs Revolutionize the Design of Explainable and Efficient TinyML Models?}, 
      author={Christophe El Zeinaty and Wassim Hamidouche and Glenn Herrou and Daniel Menard and Merouane Debbah},
      year={2025},
      eprint={2504.09685},
      archivePrefix={arXiv},
      primaryClass={cs.LG},
}

@inproceedings{cnas,
  author={Gambella, Matteo and Falcetta, Alessandro and Roveri, Manuel},
  booktitle={2022 IEEE International Conference on Systems, Man, and Cybernetics (SMC)}, 
  title={CNAS: Constrained Neural Architecture Search}, 
  year={2022},
  volume={},
  number={},
  pages={2918-2923},
  doi={10.1109/SMC53654.2022.9945080}
}

@inproceedings{gambella2023edanas,
  title={{EDANAS}: Adaptive Neural Architecture Search for Early Exit Neural Networks},
  author={Gambella, Matteo and Roveri, Manuel},
  booktitle={2023 International Joint Conference on Neural Networks (IJCNN)},
  pages={1--8},
  year={2023},
  organization={IEEE}
}

@misc{nas_survey_components,
	title = {A {Survey} on {Neural} {Architecture} {Search}},
	language = {en},
	urldate = {2023-09-01},
	publisher = {arXiv},
	author = {Wistuba, Martin and Rawat, Ambrish and Pedapati, Tejaswini},
	month = jun,
	year = {2019},
	note = {arXiv:1905.01392 [cs, stat]},
}

@article{surveydynamicNN,
  author       = {Yizeng Han and
                  Gao Huang and
                  Shiji Song and
                  Le Yang and
                  Honghui Wang and
                  Yulin Wang},
  title        = {Dynamic Neural Networks: {A} Survey},
  journal      = {CoRR},
  volume       = {abs/2102.04906},
  year         = {2021},
  eprinttype    = {arXiv},
  eprint       = {2102.04906},
  bibsource    = {dblp computer science bibliography, https://dblp.org}
}

@article{advancesnas,
    author = {Wang, Xin and Zhu, Wenwu},
    title = {Advances in neural architecture search},
    journal = {National Science Review},
    volume = {11},
    number = {8},
    pages = {nwae282},
    year = {2024},
    month = {08},
    issn = {2095-5138},
    doi = {10.1093/nsr/nwae282},
    eprint = {https://academic.oup.com/nsr/article-pdf/11/8/nwae282/59087212/nwae282.pdf},
}

@inproceedings{fbnet,
  title     = {FBNet: Hardware-Aware Efficient ConvNet Design via Differentiable Neural Architecture Search},
  author    = {Wu, Bichen and Dai, Xiaoliang and Zhang, Peizhao and Wang, Yanghan and Sun, Fei and Wu, Yuandong and Tian, Yu and Vajda, Peter and Jia, Yangqing and Keutzer, Kurt},
  booktitle = {Proceedings of the IEEE/CVF Conference on Computer Vision and Pattern Recognition (CVPR)},
  pages     = {10726--10734},
  year      = {2019},
  month     = jun,
  doi       = {10.1109/CVPR.2019.01097}
}

@inproceedings{cai2019proxylessnas,
  title     = {ProxylessNAS: Direct Neural Architecture Search on Target Task and Hardware},
  author    = {Cai, Han and Zhu, Ligeng and Han, Song},
  booktitle = {Proceedings of the International Conference on Learning Representations (ICLR)},
  pages     = {1--13},
  year      = {2019},
  note      = {arXiv:1812.00332}
}

@article{liu2023enas_survey,
  title     = {A Survey on Evolutionary Neural Architecture Search},
  author    = {Liu, Yanan and Sun, Yanan and Xue, Bing and Zhang, Mengjie and Yen, Gary G. and Tan, Kay Chen},
  journal   = {IEEE Transactions on Neural Networks and Learning Systems},
  volume    = {34},
  number    = {2},
  pages     = {550--570},
  year      = {2023},
  month     = {February},
  doi       = {10.1109/TNNLS.2021.3116717},
}

@inproceedings{dong2019nasbench201,
  title     = {Searching for a Robust Neural Architecture in Four GPU Hours},
  author    = {Dong, Xuanyi and Yang, Yi},
  booktitle = {Proceedings of the IEEE/CVF Conference on Computer Vision and Pattern Recognition (CVPR)},
  pages     = {1761--1770},
  year      = {2019},
  month     = {June},
  doi       = {10.1109/CVPR.2019.00186},
}

@inproceedings{Hutter2011SMBO,
  title     = {Sequential Model-Based Optimization for General Algorithm Configuration},
  author    = {Hutter, Frank and Hoos, Holger H. and Leyton-Brown, Kevin},
  booktitle = {International Conference on Learning and Intelligent Optimization (LION)},
  pages     = {507--523},
  year      = {2011},
  publisher = {Springer},
  doi       = {10.1007/978-3-642-25566-3_40}
}

@inproceedings{Mellor2021NASWOT,
  title        = {Neural Architecture Search without Training},
  author       = {Mellor, Joe and Turner, Jack and Storkey, Amos J. and Crowley, Elliot J.},
  booktitle    = {Proceedings of the 38th International Conference on Machine Learning (ICML)},
  year         = {2021},
}

@article{salmanipouravval2025systematic,
  author    = {Salmani Pour Avval, Sasan and Eskue, Nathan D. and Groves, Roger M. and Yaghoubi, Vahid},
  title     = {Systematic review on neural architecture search},
  journal   = {Artificial Intelligence Review},
  volume    = {58},
  pages     = {73},
  year      = {2025},
  doi       = {10.1007/s10462-024-11058-w},
}

@article{rbf,
  title={Accelerating neural architecture search using performance prediction},
  author={Baker, Bowen and Gupta, Otkrist and Raskar, Ramesh and Naik, Nikhil},
  journal={arXiv preprint arXiv:1705.10823},
  year={2017}
}

@inproceedings{mlp,
author = {Liu, Chenxi and Zoph, Barret and Neumann, Maxim and Shlens, Jonathon and Hua, Wei and Li, Li-Jia and Fei-Fei, Li and Yuille, Alan and Huang, Jonathan and Murphy, Kevin},
title = {Progressive Neural Architecture Search},
booktitle = {Proceedings of the European Conference on Computer Vision (ECCV)},
year = {2018}
}

@article{carts,
  author={Sun, Yanan and Wang, Handing and Xue, Bing and Jin, Yaochu and Yen, Gary G. and Zhang, Mengjie},
  journal={IEEE Transactions on Evolutionary Computation}, 
  title={Surrogate-Assisted Evolutionary Deep Learning Using an End-to-End Random Forest-Based Performance Predictor}, 
  year={2020},
  volume={24},
  number={2},
  pages={350-364},
  doi={10.1109/TEVC.2019.2924461}}

@InProceedings{frankwolfe,
  title = 	 {Variance-Reduced and Projection-Free Stochastic Optimization},
  author = 	 {Hazan, Elad and Luo, Haipeng},
  booktitle = 	 {Proceedings of The 33rd International Conference on Machine Learning},
  pages = 	 {1263--1271},
  year = 	 {2016},
  editor = 	 {Balcan, Maria Florina and Weinberger, Kilian Q.},
  volume = 	 {48},
  series = 	 {Proceedings of Machine Learning Research},
  address = 	 {New York, New York, USA},
  month = 	 {20--22 Jun},
  publisher =    {PMLR},
  url = 	 {https://proceedings.mlr.press/v48/hazana16.html},

}

@inproceedings{cheng2020instanas,
  title={InstaNAS: Instance-Aware Neural Architecture Search},
  author={Cheng, An-Chieh and Lin, Chieh Hubert and Juan, Da-Cheng and Wei, Wei and Sun, Min},
  booktitle={Proceedings of the AAAI Conference on Artificial Intelligence},
  volume={34},
  number={04},
  pages={3577--3584},
  year={2020}
}

@inproceedings{qosnas,
  title={QoS-aware Neural Architecture Search},
  author={Cheng, An-Chieh and Lin, Chieh Hubert and Juan, Da-Cheng and Wei, Wei and Sun, Min},
  booktitle={Advances in Neural Information Processing Systems (NeurIPS)},
  year={2019},
}

@misc{branchLLM,
      title={Towards Distributed Neural Architectures}, 
      author={Aditya Cowsik and Tianyu He and Andrey Gromov},
      year={2025},
      eprint={2506.22389},
      archivePrefix={arXiv},
      primaryClass={cs.LG},
}

@misc{odena2017changingmodelbehaviortesttime,
      title={Changing Model Behavior at Test-Time Using Reinforcement Learning}, 
      author={Augustus Odena and Dieterich Lawson and Christopher Olah},
      year={2017},
      eprint={1702.07780},
      archivePrefix={arXiv},
      primaryClass={stat.ML},
}

@INPROCEEDINGS {dynamicofa,
author = { Lou, Wei and Xun, Lei and Sabet, Amin and Bi, Jia and Hare, Jonathon and Merrett, Geoff V. },
booktitle = { 2021 IEEE/CVF Conference on Computer Vision and Pattern Recognition Workshops (CVPRW) },
title = {{ Dynamic-OFA: Runtime DNN Architecture Switching for Performance Scaling on Heterogeneous Embedded Platforms }},
year = {2021},
volume = {},
ISSN = {},
pages = {3104-3112},
doi = {10.1109/CVPRW53098.2021.00347},
publisher = {IEEE Computer Society},
address = {Los Alamitos, CA, USA},
month =Jun}

@inproceedings{manas,
author = {Chen, Hanxiong and Li, Yunqi and Zhu, He and Zhang, Yongfeng},
title = {Learn Basic Skills and Reuse: Modularized Adaptive Neural Architecture Search (MANAS)},
year = {2022},
isbn = {9781450392365},
publisher = {Association for Computing Machinery},
address = {New York, NY, USA},
doi = {10.1145/3511808.3557385},
booktitle = {Proceedings of the 31st ACM International Conference on Information \& Knowledge Management},
pages = {169–179},
numpages = {11},
location = {Atlanta, GA, USA},
series = {CIKM '22}
}

@misc{moenas,
      title={MoENAS: Mixture-of-Expert based Neural Architecture Search for jointly Accurate, Fair, and Robust Edge Deep Neural Networks}, 
      author={Lotfi Abdelkrim Mecharbat and Alberto Marchisio and Muhammad Shafique and Mohammad M. Ghassemi and Tuka Alhanai},
      year={2025},
      eprint={2502.07422},
      archivePrefix={arXiv},
      primaryClass={cs.LG},
}

@misc{cmnas,
      title={CM-NAS: Cross-Modality Neural Architecture Search for Visible-Infrared Person Re-Identification}, 
      author={Chaoyou Fu and Yibo Hu and Xiang Wu and Hailin Shi and Tao Mei and Ran He},
      year={2021},
      eprint={2101.08467},
      archivePrefix={arXiv},
      primaryClass={cs.CV},
}

@article{akiba2025evolutionary,
  title={Evolutionary optimization of model merging recipes},
  author={Akiba, Takuya and Shing, Makoto and Tang, Yujin and Sun, Qi and Ha, David},
  journal={Nature Machine Intelligence},
  volume={7},
  number={2},
  pages={195--204},
  year={2025},
  publisher={Nature Publishing Group},
  doi={10.1038/s42256-024-00975-8},
}

@article{mmnas,
  title={Deep Multimodal Neural Architecture Search},
  author={Yu, Zhou and Cui, Yuhao and Yu, Jun and Wang, Meng and Tao, Dacheng and Tian, Qi},
  journal={Proceedings of the 28th ACM International Conference on Multimedia},
  pages = {3743--3752},
  year={2020}
}

@inproceedings{bmnas,
  title     = {BM-NAS: Bilevel Multimodal Neural Architecture Search},
  author    = {Yin, Yihang and Huang, Siyu and Zhang, Xiang},
  booktitle = {Proceedings of the AAAI Conference on Artificial Intelligence},
  volume    = {36},
  number    = {9},
  pages     = {8901--8909},
  year      = {2022},
  publisher = {AAAI Press},
  doi       = {10.1609/aaai.v36i9.21245},
}

@article{feng2019learning,
  title={Learning modality-specific representations for visible-infrared person re-identification},
  author={Feng, Zhanxiang and Lai, Jianhuang and Xie, Xiaohua},
  journal={IEEE Transactions on Image Processing (TIP)},
  volume={29},
  pages={579--590},
  year={2019},
  publisher={IEEE}
}

@inproceedings{matena2022merging,
  title={Merging Models with Fisher-Weighted Averaging},
  author={Matena, Michael S. and Raffel, Colin},
  booktitle={Advances in Neural Information Processing Systems},
  volume={35},
  pages={17703--17716},
  year={2022}
}

@inproceedings{PerezRua2019MFAS,
  title     = {MFAS: Multimodal Fusion Architecture Search},
  author    = {P{\'e}rez-R{\'u}a, Juan-Manuel and Vielzeuf, Valentin and Pateux, Stéphane and Baccouche, Moez and Jurie, Frédéric},
  booktitle = {Proceedings of the IEEE/CVF Conference on Computer Vision and Pattern Recognition (CVPR)},
  pages     = {6966--6975},
  year      = {2019}
}

@article{Li2019VisualBERT,
  title     = {VisualBERT: A Simple and Performant Baseline for Vision and Language},
  author    = {Li, Liunian Harold and Yatskar, Mark and Yin, Da and Hsieh, Cho-Jui and Chang, Kai-Wei},
  journal   = {arXiv preprint arXiv:1908.03557},
  year      = {2019}
}

@misc{harmonicnas,
      title={Harmonic-NAS: Hardware-Aware Multimodal Neural Architecture Search on Resource-constrained Devices}, 
      author={Mohamed Imed Eddine Ghebriout and Halima Bouzidi and Smail Niar and Hamza Ouarnoughi},
      year={2023},
      eprint={2309.06612},
      archivePrefix={arXiv},
      primaryClass={cs.LG},
}

@misc{ersam,
      title={ERSAM: Neural Architecture Search For Energy-Efficient and Real-Time Social Ambiance Measurement}, 
      author={Chaojian Li and Wenwan Chen and Jiayi Yuan and Yingyan Celine Lin and Ashutosh Sabharwal},
      year={2025},
      eprint={2303.10727},
      archivePrefix={arXiv},
      primaryClass={cs.LG}, 
}

@INPROCEEDINGS{flatnas,
  author={Gambella, Matteo and Pittorino, Fabrizio and Roveri, Manuel},
  booktitle={2024 International Joint Conference on Neural Networks (IJCNN)}, 
  title={FlatNAS: optimizing Flatness in Neural Architecture Search for Out-of-Distribution Robustness}, 
  year={2024},
  volume={},
  number={},
  pages={1-8},
  doi={10.1109/IJCNN60899.2024.10650433}}

@inproceedings{nasood,
	address = {Montreal, QC, Canada},
	title = {{NAS}-{OoD}: {Neural} {Architecture} {Search} for {Out}-of-{Distribution} {Generalization}},
	isbn = {978-1-66542-812-5},
	shorttitle = {{NAS}-{OoD}},
	doi = {10.1109/ICCV48922.2021.00821},
	language = {en},
	urldate = {2024-01-02},
	booktitle = {2021 {IEEE}/{CVF} {International} {Conference} on {Computer} {Vision} ({ICCV})},
	publisher = {IEEE},
	author = {Bai, Haoyue and Zhou, Fengwei and Hong, Lanqing and Ye, Nanyang and Chan, S.-H. Gary and Li, Zhenguo},
	month = oct,
	year = {2021},
	pages = {8300--8309},
}

@article{tinas,
  title        = {Intermittent-Friendly Neural Architecture Search: Demystifying Accuracy and Overhead Tradeoffs},
  author       = {Hashan Roshantha Mendis and Chih-Hsuan Yen and Chih-Kai Kang and Pi-Cheng Hsiu},
  journal      = {IEEE Transactions on Computer-Aided Design of Integrated Circuits and Systems},
  volume       = {44},
  number       = {10},
  pages        = {3990--4003},
  year         = {2025},
  month        = oct,
  doi          = {10.1109/TCAD.2025.3555963}
}

@inproceedings{inas,
  author       = {Hashan Roshantha Mendis and Kas{\i}m Sinan Y{\i}ld{\i}r{\i}m and Marco Zimmerling and Luca Mottola and Pi-Cheng Hsiu},
  title        = {Intermittent TinyML: Powering Sustainable Deep Intelligence Without Batteries},
  booktitle    = {Proceedings of the International Conference on Embedded Software (EMSOFT '25)},
  year         = {2025},
  address      = {Taipei, Taiwan},
  publisher    = {ACM},
  isbn         = {979-8-4007-1993-6},
  doi          = {10.1145/3742874.3757084}
}

@article{tinymloverview,
  title={A review on TinyML: State-of-the-art and prospects},
  author={Partha Pratim Ray},
  journal={Journal of King Saud University - Computer and Information Sciences},
  year={2021}
}

@inproceedings{abbasi2020modeling,
  title={Modeling teacher-student techniques in deep neural networks for knowledge distillation},
  author={Abbasi, Sajjad and Hajabdollahi, Mohsen and Karimi, Nader and Samavi, Shadrokh},
  booktitle={2020 International Conference on Machine Vision and Image Processing (MVIP)},
  pages={1--6},
  year={2020},
  organization={IEEE}
}

@article{han2015learning,
  title={Learning both weights and connections for efficient neural network},
  author={Han, Song and Pool, Jeff and Tran, John and Dally, William},
  journal={Advances in neural information processing systems},
  volume={28},
  year={2015}
}

@article{han2015deep,
  title={Deep compression: Compressing deep neural networks with pruning, trained quantization and huffman coding},
  author={Han, Song and Mao, Huizi and Dally, William J},
  journal={arXiv preprint arXiv:1510.00149},
  year={2015}
}

@article{scalinglaw,
  author  = {Kaplan, Jared and McCandlish, Sam and Henighan, Tom and Brown, Tom B. and Chess, Benjamin and Child, Rewon and Gray, Scott and Radford, Alec and Wu, Jeffrey and Amodei, Dario},
  title   = {Scaling Laws for Neural Language Models},
  journal = {arXiv preprint arXiv:2001.08361},
  year    = {2020},
  url     = {https://arxiv.org/abs/2001.08361}
}

@misc{TCN,
	title = {An {Empirical} {Evaluation} of {Generic} {Convolutional} and {Recurrent} {Networks} for {Sequence} {Modeling}},
	url = {http://arxiv.org/abs/1803.01271},
	language = {en},
	urldate = {2023-01-24},
	publisher = {arXiv},
	author = {Bai, Shaojie and Kolter, J. Zico and Koltun, Vladlen},
	month = apr,
	year = {2018},
	note = {arXiv:1803.01271 [cs]},
}

@article{velasquez2025tinyml,
  author    = {Velasquez, Juan D. and Cadavid, Lorena and Franco, Carlos J.},
  title     = {Emerging trends and strategic opportunities in tiny machine 
               learning: A comprehensive thematic analysis},
  journal   = {Neurocomputing},
  volume    = {648},
  pages     = {130746},
  year      = {2025},
  doi       = {10.1016/j.neucom.2025.130746},
}

@ARTICLE{perceptionmodel,
  author={Zhou, Shaoze and Guo, Tianshuo and Luan, Xingsen and Li, Yonghua},
  journal={IEEE Internet of Things Journal}, 
  title={Multidimensional Edge Perception Model for Rail Vehicle Operational States Based on Artificial Intelligence of Things}, 
  year={2024},
  volume={11},
  number={18},
  pages={29728-29741},
  doi={10.1109/JIOT.2024.3405356}}

@inproceedings{munoz-etal-2024-lonas,
    title = "{L}o{NAS}: Elastic Low-Rank Adapters for Efficient Large Language Models",
    author = "Munoz, Juan Pablo  and
      Yuan, Jinjie  and
      Zheng, Yi  and
      Jain, Nilesh",
    editor = "Calzolari, Nicoletta  and
      Kan, Min-Yen  and
      Hoste, Veronique  and
      Lenci, Alessandro  and
      Sakti, Sakriani  and
      Xue, Nianwen",
    booktitle = "Proceedings of the 2024 Joint International Conference on Computational Linguistics, Language Resources and Evaluation (LREC-COLING 2024)",
    month = may,
    year = "2024",
    address = "Torino, Italia",
    publisher = "ELRA and ICCL",
    url = "https://aclanthology.org/2024.lrec-main.940",
    pages = "10760--10776",
}

@misc{compressingllm,
      title={Compressing Large Language Models with Automated Sub-Network Search}, 
      author={Rhea Sanjay Sukthanker and Benedikt Staffler and Frank Hutter and Aaron Klein},
      year={2025},
      eprint={2410.06479},
      archivePrefix={arXiv},
      primaryClass={cs.CL},
      url={https://arxiv.org/abs/2410.06479}, 
}

@misc{composer,
      title={Composer: A Search Framework for Hybrid Neural Architecture Design}, 
      author={Bilge Acun and Prasoon Sinha and Newsha Ardalani and Sangmin Bae and Alicia Golden and Chien-Yu Lin and Meghana Madhyastha and Fei Sun and Neeraja J. Yadwadkar and Carole-Jean Wu},
      year={2026},
      eprint={2510.00379},
      archivePrefix={arXiv},
      primaryClass={cs.LG},
      url={https://arxiv.org/abs/2510.00379}, 
}

@inproceedings{lora,
  title     = {LoRA: Low-Rank Adaptation of Large Language Models},
  author    = {Hu, Edward J. and Shen, Yelong and Wallis, Phillip and Allen-Zhu, Zeyuan and Li, Yuanzhi and Wang, Shean and Wang, Lu and Chen, Weizhu},
  booktitle = {International Conference on Learning Representations (ICLR)},
  year      = {2022},
  url       = {https://arxiv.org/abs/2106.09685}
}

@misc{qlora,
      title={QLoRA: Efficient Finetuning of Quantized LLMs}, 
      author={Tim Dettmers and Artidoro Pagnoni and Ari Holtzman and Luke Zettlemoyer},
      year={2023},
      eprint={2305.14314},
      archivePrefix={arXiv},
      primaryClass={cs.LG},
      url={https://arxiv.org/abs/2305.14314}, 
}

@inproceedings{dynamicsubspace,
  title     = {Subnet-Aware Dynamic Supernet Training
               for Neural Architecture Search},
  author    = {Jeon, Hoyong and others},
  booktitle = {Proceedings of the IEEE/CVF Conference on Computer
               Vision and Pattern Recognition (CVPR)},
  year      = {2025}
}

@inproceedings{lapt,
  title     = {Design Principle Transfer in Neural Architecture Search
               via Large Language Models},
  author    = {Zhou, Xun and Wang, Zhengwei and Feng, Liang
               and Liu, Songbai and Wong, Ka-Chun and Tan, Kay Chen},
  booktitle = {Proceedings of the AAAI Conference on Artificial Intelligence},
  year      = {2025},
  note      = {arXiv:2408.11330}
}

@inproceedings{llmatic,
  title     = {{LLMatic}: Neural Architecture Search via Large Language
               Models and Quality Diversity Optimization},
  author    = {Nasir, Muhammad U. and Earle, Sam and Cleghorn, Christopher
               and James, Steven and Togelius, Julian},
  booktitle = {Proceedings of the Genetic and Evolutionary Computation
               Conference (GECCO)},
  year      = {2024},
  note      = {arXiv:2306.01102}
}

@inproceedings{creamcrop,
 author = {Peng, Houwen and Du, Hao and Yu, Hongyuan and LI, QI and Liao, Jing and Fu, Jianlong},
 booktitle = {Advances in Neural Information Processing Systems},
 editor = {H. Larochelle and M. Ranzato and R. Hadsell and M.F. Balcan and H. Lin},
 pages = {17955--17964},
 publisher = {Curran Associates, Inc.},
 title = {Cream of the Crop: Distilling Prioritized Paths For One-Shot Neural Architecture Search},
 url = {https://proceedings.neurips.cc/paper_files/paper/2020/file/d072677d210ac4c03ba046120f0802ec-Paper.pdf},
 volume = {33},
 year = {2020}
}

@inproceedings{metanas,
  title     = {Meta-Learning of Neural Architectures for Few-Shot Learning},
  author    = {Elsken, Thomas and Staffler, Benedikt
               and Metzen, Jan Hendrik and Hutter, Frank},
  booktitle = {Proceedings of the IEEE/CVF Conference on Computer
               Vision and Pattern Recognition (CVPR)},
  pages     = {12365--12375},
  year      = {2020}
}

@misc{metaoptimizers,
      title={Tasks, stability, architecture, and compute: Training more effective learned optimizers, and using them to train themselves}, 
      author={Luke Metz and Niru Maheswaranathan and C. Daniel Freeman and Ben Poole and Jascha Sohl-Dickstein},
      year={2020},
      eprint={2009.11243},
      archivePrefix={arXiv},
      primaryClass={cs.LG},
      url={https://arxiv.org/abs/2009.11243}, 
}

@article{survey-CE-NAS,
title = {A survey on computationally efficient neural architecture search},
journal = {Journal of Automation and Intelligence},
volume = {1},
number = {1},
pages = {100002},
year = {2022},
issn = {2949-8554},
doi = {https://doi.org/10.1016/j.jai.2022.100002},
url = {https://www.sciencedirect.com/science/article/pii/S2949855422000028},
author = {Shiqing Liu and Haoyu Zhang and Yaochu Jin},
}

@article{surveyfederatednas,
  author    = {Zhu, Hangyu and Zhang, Haoyu and Jin, Yaochu},
  title     = {From federated learning to federated neural architecture search: a survey},
  journal   = {Complex \& Intelligent Systems},
  volume    = {7},
  number    = {2},
  pages     = {639--657},
  year      = {2021},
  publisher = {Springer},
  doi       = {10.1007/s40747-020-00247-z}
}

@Article{hapfnas,
AUTHOR = {Yang, An and Liu, Ying},
TITLE = {Heterogeneity-Aware Personalized Federated Neural Architecture Search},
JOURNAL = {Entropy},
VOLUME = {27},
YEAR = {2025},
NUMBER = {7},
ARTICLE-NUMBER = {759},
URL = {https://www.mdpi.com/1099-4300/27/7/759},
PubMedID = {40724475},
ISSN = {1099-4300},
DOI = {10.3390/e27070759}
}

@inproceedings{
benchmarking_ood,
title={Benchmarking Neural Network Robustness to Common Corruptions and Perturbations},
author={Dan Hendrycks and Thomas Dietterich},
booktitle={International Conference on Learning Representations},
year={2019}
}

@misc{squad,
      title={SQUAD: Scalable Quorum Adaptive Decisions via ensemble of early exit neural networks}, 
      author={Matteo Gambella and Fabrizio Pittorino and Giuliano Casale and Manuel Roveri},
      year={2026},
      eprint={2601.22711},
      archivePrefix={arXiv},
      primaryClass={cs.LG},
      url={https://arxiv.org/abs/2601.22711}, 
}

\end{document}